%% file: main.tex
\ifdefined\pdftexversion
  \pdfoutput=1
\fi
\documentclass[10pt, logo, onecolumn, copyright]{nv}

\input{preamble_nv}

\title{\ours{}: Hybrid Linear Attention with Attention Residuals for Efficient Video Generation}

\author{%
\parbox{\linewidth}{%
\centering
\vspace{-2pt}
{\bfseries\fontsize{9.6pt}{13.5pt}\selectfont
Junsong Chen,\quad
Jincheng Yu,\quad
Yitong Li,\quad
Shuchen Xue,\quad
Haozhe Liu,\quad
Jingyu Xin,\quad
Yuyang Zhao,\par
\vspace{0.12em}
Tian Ye,\quad
Zhangjie Wu,\quad
Zian Wang,\quad
Daquan Zhou,\quad
Ping Luo,\quad
Song Han,\quad
Enze Xie\par}
\vspace{2mm}
{\normalsize NVIDIA \par}
\vspace{0.1em}
{\normalsize
\href{https://nvlabs.github.io/Sana/Video2}{\raisebox{-0.35em}{\includegraphics[height=1.5em]{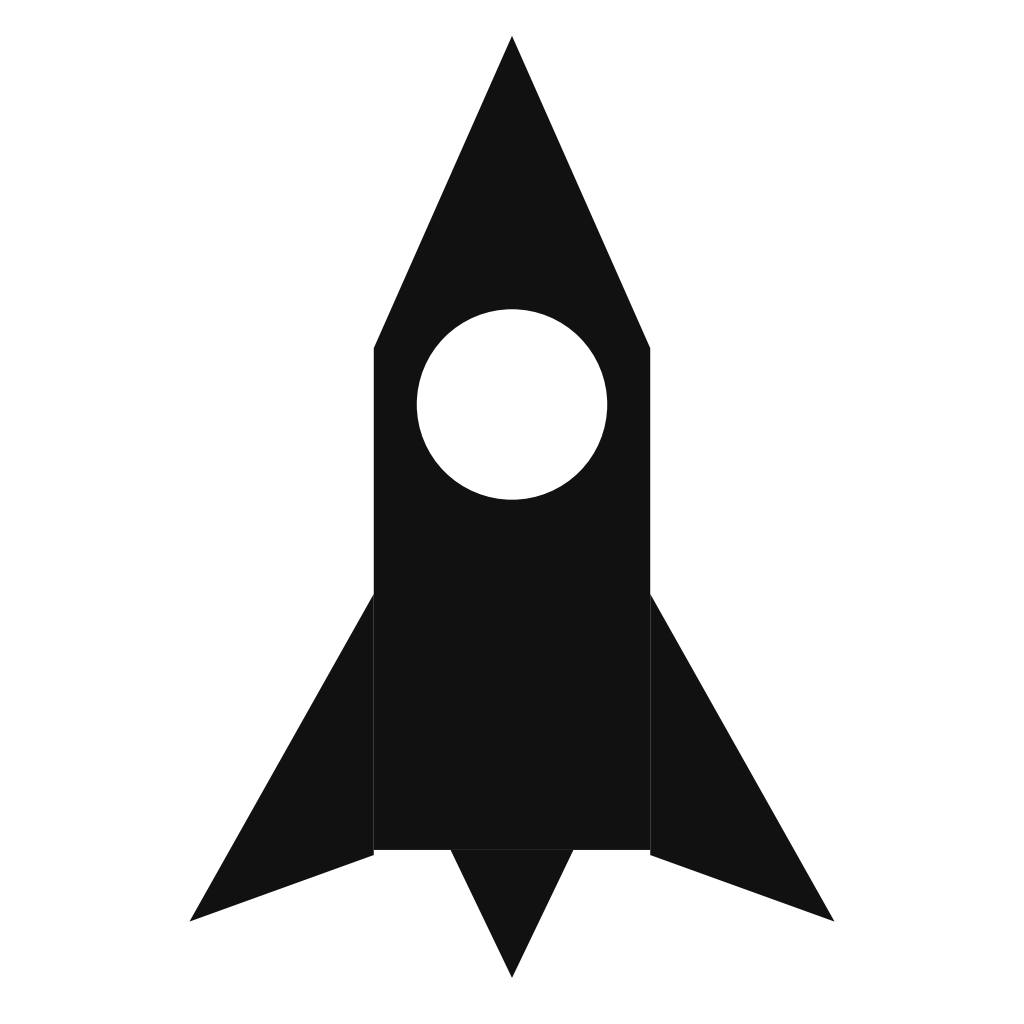}}\kern0.3em Project Page}\quad
\href{https://github.com/NVlabs/Sana}{\raisebox{-0.35em}{\includegraphics[height=1.2em,trim=256 223 256 249,clip]{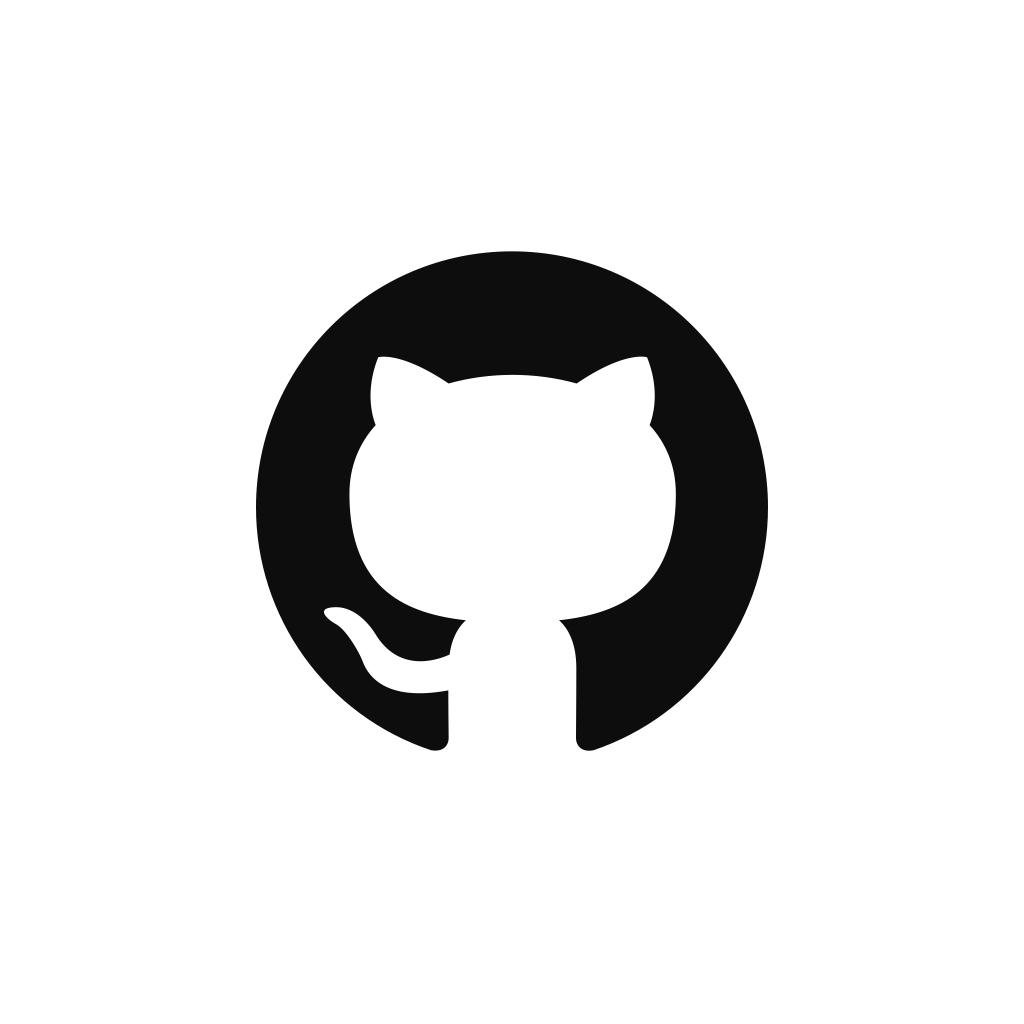}}\kern0.3em GitHub}\par}
}}

\input{sec/00_abstract}
\let\sharedpaperabstract\theabstract
\long\gdef\theabstract{\vspace*{-16pt}\noindent\textbf{Abstract:} \sharedpaperabstract}

\begin{document}

\maketitle
\enlargethispage{2cm}

\input{sec/01_introduction}
\input{sec/02_preliminary}
\input{sec/03_method}
\input{sec/04_training_recipe}
\input{sec/05_experiments}

\input{sec/06_analysis}

\input{sec/07_applications}

\input{sec/08_related_work}
\input{sec/09_conclusion}

\newpage
\input{sec/10_references}

\newpage
\appendix
\raggedbottom
\input{sec/11_appendix}

\end{document}

%% file: preamble_nv.tex
\usepackage{times}

\usepackage{nicefrac}
\usepackage{subcaption}
\usepackage{multirow}
\usepackage{wrapfig}
\usepackage{float}
\usepackage{placeins}
\usepackage[square,sort,comma,numbers]{natbib}
\usepackage{xspace}
\newcommand{\ours}{SANA-Video~2.0\xspace}
\newcommand{\prev}{SANA-Video\xspace}
\newcommand{\attnres}{AttnRes}

\newcommand{\teaserwidth}{0.95\textwidth}
\newcommand{\teasercaption}{\textbf{\ours{} at a glance.} (a)~Text-to-video examples, including embodied and physics-following scenarios. (b)~One-H100 720p/5s latency, including the final Sol-Engine 5B and 40-layer 14B results; VBench marks the 5B and Wan~2.2 quality points. (c)~5B DiT-forward speedup over duration; 14B scaling is in Appendix~\ref{app:hardware_scaling}.}

%% file: sec/00_abstract.tex
\begin{abstract}
We introduce \ours{}, a hybrid video diffusion transformer instantiated at 5B and 14B scales under a unified architecture. Designed to generate high-quality video up to 720p on a single GPU, \ours{} matches full-softmax video DiTs in quality while retaining the favorable long-sequence scaling of linear attention. To avoid quadratic attention throughout, \textbf{Hybrid Linear--Softmax Attention} combines gated linear attention for $O(N)$-dominated mixing with periodic gated-softmax anchors at a 3:1 ratio, restoring the full-rank token interactions that pure linear attention lacks. To propagate these refreshed representations across depth, \textbf{Block Attention Residuals (\attnres{})} route completed block summaries into later linear layers, enabling anchor-feature reuse and boosting deep-layer effective rank by ${\sim}12\%$. Through \textbf{from-scratch training}, \ours{} learns the complete hybrid directly rather than linearizing pretrained models, with reduced-resolution proxy studies establishing 25\% softmax as the optimal quality--efficiency trade-off. With 40-step sampling, \ours{} achieves a VBench score of $84.30$ in $13.2$s at 480p on a single H100, remaining competitive with far larger softmax video DiTs at a fraction of the latency. Its compiled DiT forward pass is \textbf{3.2$\times$ faster} than a matched full-softmax baseline at 720p/60s, a gap that expands with video duration. Furthermore, full-stack Sol-Engine optimization (kernel fusion, caching, and sparse attention) accelerates this hardware-friendly backbone by a further \textbf{3.58$\times$}, bringing the 5B pipeline to $13.06$s at 720p/5s and making it \textbf{120$\times$ faster} than Wan~2.2-A14B on one H100. Overall, our hybrid design recovers softmax-level expressiveness at substantially reduced cost, unlocking scalable long, high-resolution video generation.
\end{abstract}

%% file: sec/01_introduction.tex
\input{figure_latex/teaser}
\section{Introduction}
\label{sec:intro}

Video generation is a rapidly advancing field, powering applications from creative content production to virtual product displays and live streaming. Recent large-scale generators, including Wan~2.1~\cite{wan2025} (14B), HunyuanVideo~\cite{hunyuanvideo2024} (13B), Seedance~2.0~\cite{seedance2026seedance20advancingvideo}, and closed systems such as Veo\,3 and Kling, produce remarkably high-fidelity video. Among open Video DiT~\cite{peebles2023dit} baselines such as Wan~2.1 and HunyuanVideo, full 3D softmax attention incurs an $O(N^2)$ cost that is punishing for video: after VAE compression a single 1080p clip already spans tens of thousands of latent tokens, and in our matched production-scale profiling sweep a full-softmax backbone is $2.01\times$ slower than our 25\% hybrid at 1080p/121f, with both sides compiled on their best kernels (Figure~\ref{fig:efficiency_profiling}(a)). The problem worsens for long video ($>$10s), where the quadratic attention matrix and activation workspace dominate the cost; even dedicated long-video efforts such as MAGI-1~\cite{magi1} remain constrained by vanilla attention.

Linear attention offers a principled escape. Its $O(N)$ complexity scales gracefully with sequence length, its state-space view exposes a clean interface for recurrent kernels, and \prev{}~\cite{sanavideo2025} showed that a \emph{pure}-linear Video DiT already delivers large speedups at competitive quality. But pure linear attention pays for its efficiency with expressiveness: the fixed-size state matrix $S \in \mathbb{R}^{d \times d}$ cannot represent every token--token interaction, which can weaken precise spatiotemporal correspondence and fine detail~\cite{zhang2024rala,arora2024based}. Recent large language models point to a fix: keep a mostly linear-attention stack but insert a few softmax \emph{anchors}---periodic full-attention layers that restore exact, less rank-constrained token interactions at fixed depths---at a regular ratio (Qwen3-Next~\cite{qwen3next}, Kimi-Linear~\cite{yang2025kimi}), and route information across depth with Attention Residuals~\cite{kimiteam2026attnres}, a combination the recent Kimi~K3~\cite{kimik3} adopts at trillion-parameter scale. This raises a central question: \emph{can a mostly-linear backbone recover the expressiveness of softmax attention while keeping its $O(N)$ scaling for long, high-resolution video?}

This paper answers affirmatively with \ours{}, a hybrid-attention video diffusion transformer instantiated at 5B and 14B scales. The larger model is a 40-layer, width-4,096 backbone with 14.25B parameters, trained at 384$\times$B200 scale. The 5B operating point keeps a mostly-linear backbone but inserts a small number of softmax anchors, matching strong full-softmax Video DiTs in quality while remaining fast on a single GPU (Figure~\ref{fig:teaser}): with 40-step sampling it reaches VBench Total $84.30$ in $13.2$s at $480{\times}832{\times}81$ on one H100, competitive with much larger softmax models at a fraction of their latency, and its DiT forward is $3.2\times$ faster than a matched full-softmax DiT at a 720p/60s shape with both compiled on their best kernels (Figure~\ref{fig:teaser}(c)), an advantage that widens with duration. Unlike post-hoc linearization of a pretrained softmax model, we train the hybrid backbone directly. The design rests on three key components.

\textbf{Hybrid Linear-Softmax Attention.} We keep gated bilinear linear attention as the core token-mixing operation for its $O(N)$ cost, and interleave gated softmax anchors at a regular 3:1 ratio (25\% softmax) placed at fixed depths. The anchors periodically restore the less rank-constrained token interactions that pure linear attention cannot express, while the linear majority preserves long-sequence scaling. Text enters through cross-attention, and a convolution-free SwiGLU FFN replaces \prev{}'s temporal-convolution FFN, whose measured overhead grows with duration (Appendix~\ref{app:temporal_conv}), keeping the backbone easy to profile and fuse with off-the-shelf kernels.

\textbf{Block Attention Residuals (\attnres{}).} Additive residuals alone force later layers to re-derive information computed upstream. Instead of this re-derivation, \attnres{} routes completed block-feature summaries across depth, so later linear layers can reuse the anchors' rank-refreshed updates. In a controlled same-checkpoint on/off probe, enabling \attnres{} raises deep-layer state effective rank by ${\sim}12\%$. We adapt the routing to bidirectional video diffusion by sharing routing queries across depth and removing explicit diffusion-timestep conditioning, made redundant by AdaLN.

\textbf{From-Scratch Training and Design.} We select the architecture with short, reduced-resolution proxy studies that identify 25\% softmax anchors as a practical quality--efficiency knee. The full model is then trained from scratch through a complete, documented multi-stage pipeline, comprising diverse-source data curation and filtering, a low-to-high resolution and duration curriculum, structured captioning, and self-distillation, followed by preference-based post-training (DPO and ReFL), so the hybrid learns motion and appearance without relying on a pretrained softmax or image prior. We report this training and inference-optimization pipeline in full as a central, reproducible contribution.

In conclusion, \ours{} attains competitive VBench quality ($84.30$) at markedly lower latency than large softmax Video DiTs, and its $O(N)$-dominated backbone scales far better with duration. A separate 50-step Sol-Engine deployment measures a $3.58\times$ end-to-end speedup on B200, and quantization-aware training matches BF16 quality at MXFP4 weights and MXFP8 activations (Section~\ref{sec:applications}). Rank and routing analyses reveal a clear division of labor, where linear layers provide inexpensive global mixing, softmax anchors inject exact, less rank-constrained interactions, and \attnres{} carries block-level features across depth. We hope \ours{} offers a practical, efficient foundation for high-quality video generation that researchers and everyday users can run fast.

%% file: figure_latex/teaser.tex
\providecommand{\teaserwidth}{\textwidth}
\providecommand{\teasercaption}{\textbf{\ours{} at a glance.} (a)~Text-to-video examples, including embodied and physics-following scenarios. (b)~One-H100 720p/5s latency, including the final Sol-Engine 5B and 40-layer 14B results; VBench marks the 5B and Wan~2.2 quality points. (c)~5B DiT-forward speedup over duration; 14B scaling is in Appendix~\ref{app:hardware_scaling}.}
\vspace{-6pt}
\begin{center}
\includegraphics[width=\teaserwidth]{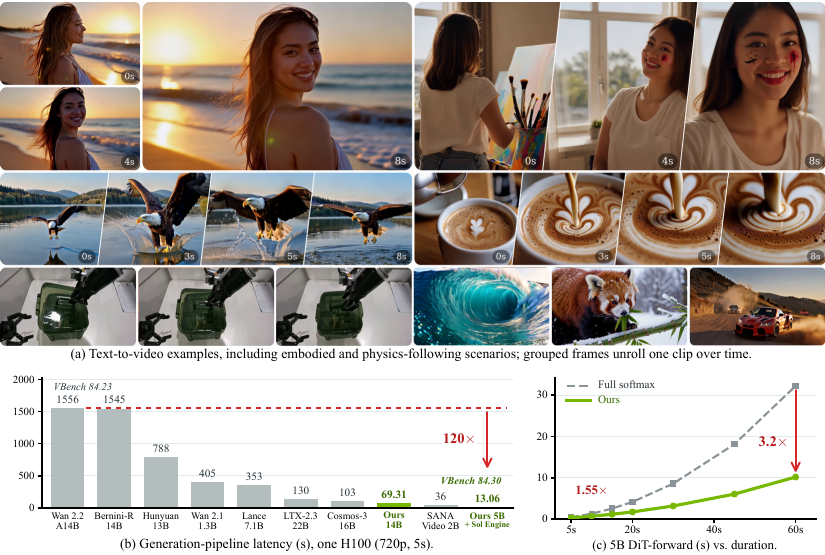}
\vspace{-5pt}
\captionof{figure}{\teasercaption}
\label{fig:teaser}
\end{center}
\newpage

%% file: sec/02_preliminary.tex
\section{Preliminaries}
\label{sec:prelim}

\paragraph{Video DiTs and flow matching.}
Video Diffusion Transformers~\cite{peebles2023dit} can be trained with conditional optimal-transport flow matching~\cite{lipman2023flow}: they form a latent sequence $z_t = (1{-}t)z + t\epsilon$ ($\epsilon \sim \mathcal{N}(0,I)$, $t \sim$ logit-normal) and predict the conditional velocity $\epsilon - z$ with loss $\mathcal{L} = \mathbb{E}\|v_\theta - (\epsilon{-}z)\|^2$. Here $t \in [0,1]$ denotes the normalized flow-matching timestep. Each block uses AdaLN-style timestep modulation, self-attention, text cross-attention, and a FFN. At 1080p a long clip already spans tens of thousands of latent sites, making softmax's $O(N^2)$ cost prohibitive.

\paragraph{Gated linear attention.}
Linear attention compresses token interactions into a fixed-size state, reducing sequence complexity to $O(Nd^2)$ at the cost of a rank bottleneck. Our bidirectional operator is a gated bilinear form rather than a causal delta-rule recurrence; dropping the delta-rule update also makes it a natural initialization for a causal Gated DeltaNet, a direction we leave to future work. For normalized queries and keys, let $q_j^r,k_n^r$ denote their RoPE-rotated forms and $\phi(\cdot)=\operatorname{ReLU}(\cdot)$. Each head computes
\begin{equation}
S=\sum_n v_n(\beta_n k_n^r)^\top,\qquad
o_j=W_o\!\left[\operatorname{RMSNorm}\!\left(\frac{S q_j^r}{\sum_n \phi(k_n)^\top\phi(q_j)+\epsilon}\right)\odot\sigma(g_j)\right],
\end{equation}
where $\beta$ gates writes to the state and $g$ gates the output. Periodic softmax layers provide less rank-constrained token mixing at selected depths while leaving most sequence computation linear. These anchors use bidirectional SDPA with QK normalization, RoPE, and the sigmoid output gate studied by Qiu et al.~\cite{qiu2025gated}. We adopt the regular 3:1 hybrid layout of Qwen3-Next~\cite{qwen3next}; Kimi-Linear~\cite{yang2025kimi} independently uses the same proportion with a~different~linear~operator.

\paragraph{Attention Residuals.}
Standard additive residuals $h_l=h_{l-1}+f_l(h_{l-1})$ propagate information through every intervening depth. Attention Residuals~\cite{kimiteam2026attnres} instead use learned depth-wise aggregation, $h_l=\sum_i\alpha_{i\to l}v_i$. Their Block AttnRes variant keeps completed block summaries and a running within-block residual, and applies separate routing before attention and FFN sublayers. In our hybrid stack, completed summaries include softmax-anchor updates and expose them to later mostly-linear layers. The original parameterizes routing queries per sublayer. Our video adaptation instead shares them across depth. Rather than assume the routing queries need a separate diffusion-timestep input, we test whether one is necessary and find it redundant with AdaLN.

%% file: sec/03_method.tex
\section{\ours{}}
\label{sec:method}

\subsection{Overview}
\label{sec:arch_overview}

\ours{} is a video DiT that combines hybrid sequence attention with block residual attention across depth. It operates on LTX-VAE~2.3~\cite{ltxvideo} latents and draws text features from Gemma-2-2B-IT~\cite{gemma2} through cross-attention at every layer. At each depth, \attnres{} first routes a representation into the self-/cross-attention branch and, in a second step, into the SwiGLU FFN, with AdaLN-style modulation inside each branch, and a final aggregation precedes the output head. The local path is convolution-free throughout. Appendix~\ref{app:config} lists the full configuration.

\input{figure_latex/architecture}

\subsection{Hybrid Attention Design}
\label{sec:hybrid}

Building on the gated attention primitives (Section~\ref{sec:prelim}), we interleave bidirectional gated linear attention layers with gated softmax anchors at a 3:1 ratio, 75\% linear, 25\% softmax, placing a softmax anchor at every fourth layer. The linear majority keeps token mixing at $O(N)$, while the uniformly spaced anchors periodically refresh the less rank-constrained interactions that the compressed linear state cannot represent (Section~\ref{sec:rank_analysis}). Linear and softmax heads use different head dimensions, trading efficiency against per-head capacity (Appendix~\ref{app:config}). This regular layout follows the hybrid linear attention of recent LLMs (Qwen3-Next~\cite{qwen3next}, Kimi-Linear~\cite{yang2025kimi}, and Kimi~K3~\cite{kimik3} at trillion-parameter scale) and maps cleanly onto fused linear-attention kernels. We confirm the $25\%$ anchor ratio as a quality--efficiency knee for the video regime by sweeping it from scratch rather than assuming the language-model value (Section~\ref{sec:softmax_ablation}). The two paths retain separate RoPE~\cite{su2024roformer} tensors and gating parameterizations: linear layers use both the write gate $\beta$ and an output gate, whereas softmax anchors use the sigmoid output gate of Qiu et al.~\cite{qiu2025gated}.

\subsection{Block Attention Residuals (\attnres{})}
\label{sec:attnres}

Figure~\ref{fig:architecture} summarizes the hybrid layer and its block-level depth routing. Hybrid attention supplies less rank-constrained token mixing only at sparse depths; \attnres{} complements it by exposing completed block summaries containing those updates to later layers. We build on the Block AttnRes~\cite{kimiteam2026attnres}, which Kimi~K3~\cite{kimik3} pairs with hybrid linear attention at language-model scale, and adapt it to bidirectional video diffusion. The layers are grouped into consecutive blocks: each block that has finished by layer $l$ leaves a single feature summary, and the block still in progress carries a running partial sum. The router at layer $l$ therefore draws on a source set $\mathcal{V}_l=\{b_0,\,b_1,\dots,\,p_l\}$ made of the initial token embedding $b_0$, the completed block summaries $b_1,b_2,\dots$ from blocks that finished before layer $l$, and the current within-block partial sum $p_l$, the accumulation so far in the in-progress block (absent only at a block boundary, where it has just been frozen into a new summary and reset). For each video token $x$, the routed representation before sublayer type $\tau$ is
\begin{equation}
  h_l(x) = \sum_{v_i\in\mathcal{V}_l} \alpha^{(\tau)}_{i\to l}(x)\, v_i(x),
  \qquad
  \alpha^{(\tau)}_{i\to l}(x) = \operatorname*{softmax}_i\!\Big( \big(w^{(\tau)} + \phi_\tau(t)\big)^{\!\top} \operatorname{RMSNorm}(v_i(x)) \Big),
  \label{eq:attnres}
\end{equation}
where $v_i(x)$ is the feature that source $v_i\in\mathcal{V}_l$ holds at token position $x$, $w^{(\tau)}$ is shared by all depths for $\tau\in\{\mathrm{attn},\mathrm{ffn}\}$, and $\phi_\tau(t)$ is an optional, zero-initialized timestep offset. The softmax is over depth sources independently for every token. After the last block, a third shared query aggregates the initial embedding and completed block sums into the final representation passed to the output head.

\vspace{-1em}
\paragraph{Shared routing queries.}
The original AttnRes learns a separate routing query for every residual sublayer, a per-layer design that language models keep even at scale (e.g., Kimi~K3~\cite{kimik3}). For our video model, one shared routing query per branch suffices: a single attention query and a single FFN query, reused at every depth. This matches per-layer routing in loss at a fraction of the memory (Section~\ref{sec:attnres_ablation}). Sharing the query does not make routing uniform. The source set $\mathcal{V}_l$ changes with depth, so the routing weights still vary from layer to layer, and because each source is normalized per token, they also vary from token to token. Sharing thus drops the per-depth query parameters while still routing the attention and FFN branches differently.

\vspace{-1em}
\paragraph{Block organization.}
We group every $S{=}8$ consecutive transformer layers into one block. As the forward pass moves through a block, each attention or FFN sublayer's output, the residual update it would otherwise add to the hidden state, is added into a running partial sum $p_l$, which the router also reads back as a source (Eq.~\ref{eq:attnres}). When the block ends, $p_l$ is frozen as that block's completed summary $b_k$, and the next block starts a fresh partial sum. Because the router keeps only one summary per finished block rather than one feature per layer, the stored history shrinks from $O(NLd)$ to $O(N\lceil L/S\rceil d)$ for sequence length $N$, roughly an $S$-fold reduction. The span of $S{=}8$ covers two softmax-anchor cycles while keeping this history compact, and the ablation treats it as an engineering default rather than a universal optimum (Section~\ref{sec:attnres_ablation}). As in Block AttnRes, attention and FFN routers learn distinct preferences (Section~\ref{sec:borrows}).

\paragraph{Timestep-independent final design.}
Although denoising timesteps have different semantic roles, the learned offset is nearly constant and changing or shuffling it has negligible validation effect. The final model therefore uses $\phi_\tau\equiv0$ while retaining timestep modulation in the DiT blocks (Section~\ref{sec:attnres_ablation}, Appendix~\ref{app:tcond_probe}).

%% file: figure_latex/architecture.tex
\begin{figure*}[t]
    \centering
    \includegraphics[width=\textwidth]{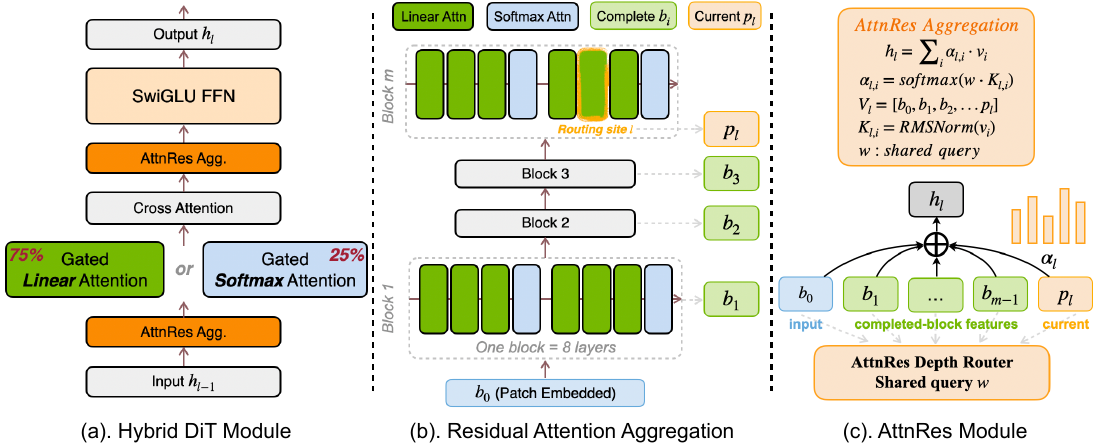}
    \vspace{-4pt}
    \caption{\looseness=-1 \textbf{\ours{} overview.} (a)~Hybrid DiT layer. (b)~Eight-layer blocks expose completed features across depth. (c)~A shared-query \attnres{} router aggregates the input and completed block features.}
    \label{fig:architecture}
    \vspace{-6pt}
\end{figure*}

%% file: sec/04_training_recipe.tex
\section{Training Recipe}
\label{sec:training_recipe}

\ours{} uses 5B and 14B models with the same hybrid design. The 5B production recipe spans data curation, a resolution/duration curriculum, and preference-based post-training (Figure~\ref{fig:training_pipeline}). The 40-layer 14B run uses the same core flow-matching and hybrid-attention formulation with scale-specific pre-training settings on 384$\times$B200 GPUs. Beyond the architecture, the recipe specifies how the corpus is transformed into progressively cleaner supervision, how noise levels are sampled, and how resolution and duration are increased. We describe these components because they are part of the method. The recipe was selected jointly rather than component-wise ablated. Appendix~\ref{app:config} lists the scale-specific architecture and training settings.

\input{figure_latex/training_pipeline}

\vspace{-1em}
\subsection{Data Curation and a Quality/Motion Funnel}
\label{sec:data_funnel}
Starting from the raw video corpus, we apply a uniform processing pipeline: shot segmentation, black-bar and subtitle cleanup, removal of low-resolution/blurry/static/exposure-defective clips, and multi-axis scoring rather than a single aggregate quality score. Keeping the axes separate is deliberate: it prevents selection from collapsing to visually clean but nearly static clips while still retaining high-quality low-motion content. Appendix~\ref{app:data_curation} gives the complete six-stage pipeline and scoring signals (Table~\ref{tab:scoring_metrics}). The resulting clips form a progressive quality/motion funnel: pre-training uses a broad pool under permissive thresholds, continual training raises the quality and motion floors, and SFT concentrates on a small trusted subset. Captions progress from mixed short/long descriptions to dense structured ones, and quantized motion descriptors are appended when available to turn the selection signal into a conditioning cue. The same quality and motion axes also drive the pre-training timestep policy below, establishing broad coverage early and focusing later stages on fidelity and prompt adherence.

\vspace{-1em}
\subsection{Training Objectives and Timestep Sampling}
\label{sec:obj_timestep}
Training uses flow matching with a logit-normal timestep density. In later continual and SFT stages we tail-floor this density, reserving probability mass near the clean and pure-noise extremes that a plain logit-normal undersamples. These regimes respectively correspond to final refinement and global structure. From the continual stage onward we make flow-shift token-count-aware: rather than a fixed per-resolution constant, we anchor the shift at two endpoints and interpolate in log-shift space by latent token count, running log-linearly from $3$ at $4{,}290$ latent tokens (480p, 81~frames) to $6$ at $23{,}000$ tokens (720p, 193~frames). Each shift is realized by offsetting the (tail-floored, $\sigma{=}0.95$) logit-normal mean by $\log(\text{shift})$, so every intermediate resolution$\times$duration lands on a principled shift and more probability mass moves to high noise as clips lengthen (Table~\ref{tab:training_curriculum}). Since the curriculum varies resolution and duration jointly, token count spans a continuous range that discrete per-resolution shifts miss. Each rank therefore computes its shift directly from the local latent shape; because production uses batch size one per rank, the mapping is per-sample.

\paragraph{Content-aware flow-shift (TQD).}
High-quality video data is scarce, so we want each clip to contribute where it teaches best. This runs into a motion--quality dilemma: motion-rich clips often carry artifacts that lower their per-frame aesthetics, whereas the cleanest clips tend to be nearly static, so a single quality bar sacrifices either motion or fidelity. TQD~\cite{tqd} resolves this by routing each clip to the noise regime where its content matters and its weaknesses do not: high noise governs global structure and motion and masks fine texture, so it receives motion-rich clips, while low noise governs fine detail and receives aesthetically clean, low-motion clips. In pre-training, clips that pass only the high-motion threshold receive a $+1.1$ logit offset, clips that pass only the high-quality threshold receive a $-1.1$ offset, and clips satisfying both or neither keep the base density. From continual training onward the TQD offset is disabled and the token-count shift takes over. Its minimum shift of~3 approximately matches the median of the high-motion TQD branch, but the change is a stage-level replacement rather than a per-sample continuous handoff. Appendix~\ref{app:tqd} gives the bias magnitudes, thresholds, and full sampling densities.

\vspace{-1em}
\paragraph{Weight EMA and Self-Flow.}
We keep an exponential moving average (EMA) of the model weights active throughout training, including SFT. Alongside it, pre-training and continual training additionally carry Self-Flow~\cite{selfflow2026}, an auxiliary feature-distillation objective with within-clip dual-timestep conditioning that we use as a training aid rather than a studied component. Self-Flow (Appendix~\ref{app:selfflow}) leaves every latent token supervised and is turned off for SFT while the weight EMA stays on.

\input{table_latex/training_curriculum}

\subsection{Timestep-Stratified Validation}
\label{sec:timestep_validation}
During training we track checkpoint quality by validation loss, but a single random-$t$ mean is a weak monitor: because loss varies several-fold across the noise axis, one scalar blurs which noise regimes improve or regress and carries high sampling variance. We therefore stratify validation into ten equal 100-step noise buckets, which exposes structure the mean hides. Across four paired checkpoints the bucket macro falls by 6.42\%, but that net figure separates a large 11.44\% low-noise improvement from a small 1.16\% high-noise regression (Figure~\ref{fig:timestep_bucket_trajectory}). Stratifying the fixed 100-evaluation budget also cuts the standard deviation of this early-to-final estimate by $2.27\times$ relative to IID timestep draws, giving a more reliable monitor. VBench Total rising from $82.68$ to $83.29$ is consistent with this net improvement, read descriptively given only four points. Appendix~\ref{app:timestep_buckets} details the matched-sample protocol and uncertainty analysis.

\subsection{Post-Training Stage}
\label{sec:post_training}

\subsubsection{Direct Preference Optimization}
\label{sec:post_training_dpo}

We employ Diffusion-DPO~\cite{wallace2024diffusion} as a standard post-training step for visual-preference alignment. For each prompt $c$, Gemini ranks videos sampled with different random seeds by visual fidelity, prompt alignment, motion quality, and temporal coherence. After filtering unreliable rankings, the highest- and lowest-ranked videos form a preferred--rejected pair $(x^+,x^-,c)$. This ranking is performed offline; Gemini is not part of the training loop.

We initialize both the trainable policy $\theta$ and a frozen reference model $\theta_{\mathrm{ref}}$ from the SFT checkpoint. At each training step, the two videos in a pair share the same sampled timestep $t$ and Gaussian noise $\epsilon$. Let $q_t$ denote the forward noising operator and $v_\phi$ the flow-velocity prediction of model $\phi$. The per-video flow-matching errors are
\begin{equation}
  x_t^\pm = q_t(x^\pm,\epsilon), \qquad
  u^\pm = \epsilon-x^\pm, \qquad
  e_\phi^\pm = \frac{1}{D}\left\lVert v_\phi(x_t^\pm,t,c)-u^\pm \right\rVert_2^2,
  \label{eq:dpo_flow_error}
\end{equation}
where $D$ is the number of latent elements. Sharing $t$ and $\epsilon$ ensures that pairwise differences reflect the videos rather than mismatched corruption levels. We define the preferred--rejected error gap as $\Delta_\phi=e_\phi^- - e_\phi^+$ and optimize
\begin{equation}
  \mathcal{L} = -\mathbb{E}\!\left[w\log\sigma\!\left(\beta(\Delta_\theta-\Delta_{\theta_{\mathrm{ref}}})\right)\right]
  +
  \lambda\,\mathbb{E}[e_\theta^+].
  \label{eq:dpo_objective}
\end{equation}
The first term is the Diffusion-DPO loss: $w$ is the pair weight derived from the judge score gap and $\beta$ controls the preference strength. The second term is a preferred-sample flow-matching regularizer that stabilizes post-training. Only the policy is updated; the reference model remains fixed throughout.

\subsubsection{Online Reinforcement Learning}
\label{sec:post_training_online_rl}

We further align the model through online reward optimization using Reward Feedback Learning (ReFL)~\cite{xu2023imagereward}. For each text prompt, ReFL first rolls out the denoising trajectory without gradients to a randomly sampled update step. At that state, our diffusion model predicts the clean latent $\hat{x}_0$, which is decoded into a video. We select the first, middle, and last frames of the decoded video and evaluate them with frozen image reward models, using the resulting frame-level feedback to optimize the video. Gradients from the rewards are propagated through the decoder and the model evaluation, avoiding backpropagation through the complete sampling trajectory.

Our joint reward combines HPSv3++~\cite{hpsv3pp}, DeQA-Score~\cite{deqa_score}, and UniPercept~\cite{unipercept}. The three models provide complementary supervision. HPSv3++ targets wide-spectrum human preference alignment across the capability--iteration spectrum, with particular emphasis on \emph{text fidelity} and \emph{aesthetic quality}. DeQA-Score provides a no-reference perceptual-quality signal: it models continuous human quality judgments through a predicted \emph{score distribution}, making it sensitive to fidelity loss and visible image degradations. UniPercept supplies a finer-grained perceptual profile spanning Image Aesthetics Assessment (IAA), Image Quality Assessment (IQA), and Image Structure and Texture Assessment (ISTA). After normalization and capping, we combine the three reward signals with an HPSv3++:DeQA-Score:UniPercept weight ratio of $4{:}4{:}1$. We additionally apply a velocity-space mean-squared-error regularizer against the frozen base model, balancing preference optimization against preservation of the generative prior acquired during pre-training and supervised fine-tuning. Section~\ref{sec:post_training_results} evaluates this online stage.

%% file: figure_latex/training_pipeline.tex
\begin{figure}[t]
\centering
\includegraphics[width=0.95\textwidth]{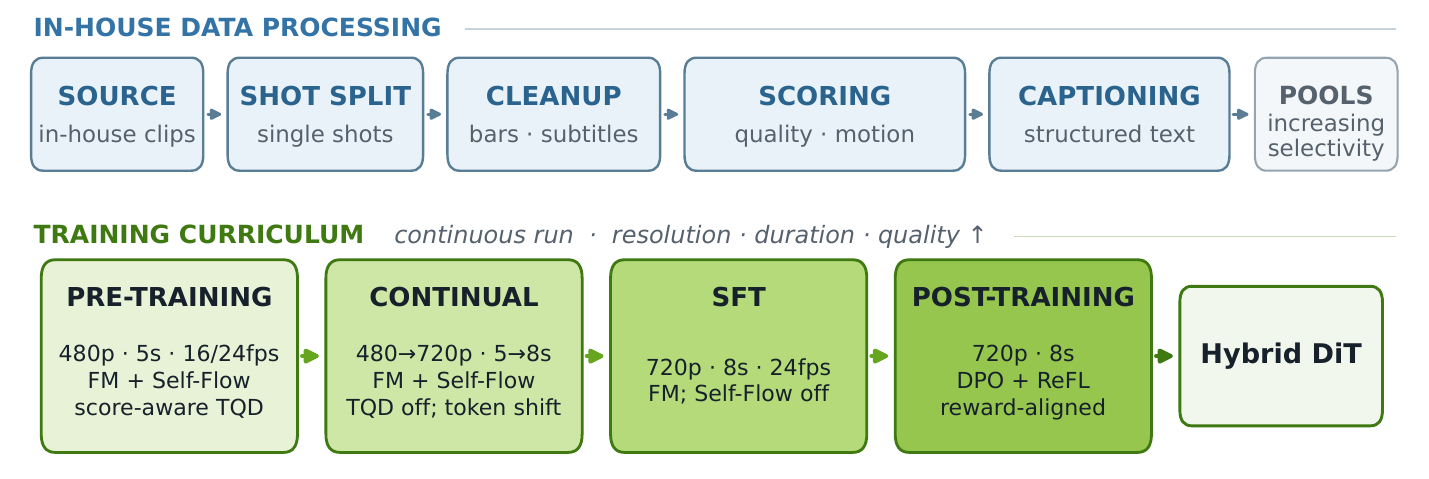}
\caption{\looseness=-1 \textbf{Training and data pipeline for the 5B checkpoint.} Curated in-house clips feed pre-training, continual training, SFT, and preference-based post-training (DPO + ReFL), with stage-specific data selection, video shapes, and training objectives per stage.}
\label{fig:training_pipeline}
\vspace{-1em}
\end{figure}

%% file: table_latex/training_curriculum.tex
\begin{table}[!t]
\centering
\caption{\looseness=-1 \textbf{Production training curriculum.} Clip counts are reported after filtering and objective add-ons are stage-specific; scale-specific model and optimizer settings are listed in Table~\ref{tab:model_config}.}
\label{tab:training_curriculum}
\small
\setlength{\tabcolsep}{5pt}
\resizebox{0.8\textwidth}{!}{%
\begin{tabular}{@{}lccccccc@{}}
\toprule
Stage & Clips & Resolution & FPS & Duration & Flow-shift & LR & Objective add-on \\
\midrule
Pre-train & ${\sim}30$M & 480p & 16/24 & 5s & 1 (+TQD) & $10^{-4}$ & TQD + Self-Flow \\
Continual & ${\sim}10$M & 480$\to$720p & 16/24 & 5s$\to$8s & 3--6 (by tokens) & $10^{-4}$ & Self-Flow \\
SFT & ${\sim}10^{4}$ & 720p & 24 & 8s & 3--6 (by tokens) & $5{\times}10^{-5}$ & standard FM \\
\bottomrule
\end{tabular}
}
\end{table}

%% file: sec/05_experiments.tex
\section{Experiments}
\label{sec:experiments}

\subsection{Implementation Details}
\label{sec:setup}

\ours{} has 5B and 14B configurations of the same hybrid backbone. Table~\ref{tab:model_config} lists both architectures and their scale-specific training settings. The 5B configuration uses 32 layers at width 2,560; the 14B configuration uses 40 layers at width 4,096 (14.25B parameters) and is trained at 384$\times$B200 scale. Both use LTX-VAE~2.3 latents, Gemma text features, flow matching, 25\% softmax anchors, and Block \attnres{}. To select the architecture and diagnose mechanisms rather than estimate final quality, we use short, reduced-resolution studies: the attention-ratio and early \attnres{} sweeps use depth-28, width-3,072 architecture-search backbones at 256p, while the query/timestep ablations and mechanism probes use 5B architecture-family checkpoints (all on 8$\times$H100 GPUs).

\input{table_latex/main_results}

\FloatBarrier
\input{figure_latex/softmax_ablation}

\subsection{Main Results}
\label{sec:main_results}

Table~\ref{tab:main_results} compares \ours{} with state-of-the-art video generators on VBench~\cite{vbench}, while Figure~\ref{fig:teaser}(b) compares complete one-H100 generation-pipeline latency, including both our 5B and 40-layer 14B configurations under the same 40-step recipe. At its $480{\times}832{\times}81$ operating point, \ours{} reaches VBench Total $84.30$ with the best Quality in Table~\ref{tab:main_results} ($85.61$). Only the author-reported Bernini-R 14B~\cite{bernini2026} scores higher on Total and Semantic ($84.64/82.49$ vs.\ $84.30/79.05$), yet our 5B model is $31.8\times$ cheaper to run under a matched shape, step count, and device ($13.2$s vs.\ $421$s on one H100). Its longer 121- and 193-frame configurations reach Total $85.29$ and $84.48$ (Appendix Table~\ref{tab:vbench_full}), and at a 720p-class shape it stays $3$--$4\times$ faster than Cosmos-3 Nano and LTX-2.3 and roughly $50\times$ faster than Bernini-R and Wan~2.2-A14B. The 40-layer, width-4,096 14B configuration takes $29.1$s at $480{\times}832{\times}81$, $14.5\times$ faster than matched Bernini-R. Full per-shape timings and the complete 16-dimension breakdown are in Appendix~\ref{app:vbench}; controlled 14B backbone scaling is reported separately in Appendix~\ref{app:hardware_scaling}.

\subsection{Design Space Exploration}
\label{sec:ablations}

We first explore the sequence-attention ratio with depth-28, width-3,072 architecture-search proxies, trained from scratch at 256p/81 frames. The sweep fixes the backbone dimensions, patch size, attention head dimensions, data, optimizer settings, seed, and training recipe while changing the softmax-layer placement; we report training curves and validation loss on 200 held-out videos with text conditioning at 10K steps. Because validation loss is strongly non-uniform across the noise axis (Figure~\ref{fig:loss_profile}), we read every proxy comparison per timestep rather than as a single scalar. The subsequent \attnres{} probes follow the per-panel protocols in Table~\ref{tab:attnres_ablation}; comparisons are made only within each matched panel.

\vspace{-1em}
\subsubsection{Softmax Ratio}
\label{sec:softmax_ablation}
We train five variants spanning 0\% (all-linear) to 100\% (all-softmax), with three hybrid ratios in between (Figure~\ref{fig:softmax_ablation}). The lowest-ratio hybrid uses four periodic softmax anchors in the 28-layer proxy ($4/28{=}14.3\%$).

\noindent\textbf{Finding:} Hybrid ratios perform better than the all-linear and all-softmax endpoints in this fixed-backbone-dimension proxy search. Pure linear attention is the weakest endpoint (all-linear val loss 0.955), consistent with the absence of periodic softmax correction. All-softmax is also worse than the hybrids (0.945), with the two endpoints close and the linear one slightly worse. Among hybrids, 50\% achieves the lowest loss (0.897) but at steep efficiency cost (1.29$\times$ latency over 25\% hybrid at 1080p/121f; Figure~\ref{fig:efficiency_profiling}(a)), while 25\% (0.905) and the four-anchor (14.3\%) hybrid (0.914) trail by only ${\sim}1\%$. Together, Figures~\ref{fig:softmax_ablation} and~\ref{fig:efficiency_profiling}(a) identify 25\% softmax as a practical Pareto knee rather than the minimum-loss point. Per-timestep analysis shows the hybrid--endpoint differentiation concentrates at mid noise ($t\approx0.2$--$0.5$).

\noindent\textit{Lock-in:} \textbf{25\% softmax (3:1 ratio)} for all subsequent experiments. This is the measured proxy Pareto knee, aligns with \attnres{} block boundaries ($S{=}8$), and matches the Qwen3-Next layout~\cite{qwen3next}; Kimi-Linear~\cite{yang2025kimi} independently uses the same proportion with a~different~operator.

\noindent\textbf{Scratch-training stability.}
All curves in Figure~\ref{fig:loss_curves} are trained from scratch; the hybrid variants converge smoothly with no loss spike or collapse and continue into the production training curriculum. The 3:1 ratio is therefore learned directly rather than obtained by post-hoc linearization.

\FloatBarrier
\subsubsection{Residual Attention}
\label{sec:attnres_ablation}
With the 3:1 hybrid ratio fixed, we study \attnres{} and select its router design (Table~\ref{tab:attnres_ablation}).

\noindent\textbf{Router across training (Table~\ref{tab:attnres_ablation}a,b).}
During early training, short-run probes select the router formulation: removing its explicit timestep input improves loss from 0.962 to 0.920. After the hybrid backbone has matured, a continued-training comparison finds \attnres{} and its removal reach comparable held-out MSE ($0.4851$ vs.\ $0.4855$, with 17 of 20 noise buckets slightly favoring \attnres{}). We do not read a quality advantage from this narrow margin; we adopt \attnres{} for the cross-depth reuse it provides and use a timestep-independent router, verifying the intended behavior through the rank and routing probes in Figure~\ref{fig:attnres_evidence}.

\noindent\textbf{Query design (Table~\ref{tab:attnres_ablation}c).}
Per-layer and shared projections have nearly identical loss (0.496 vs.\ 0.495), but sharing reduces memory overhead $4\times$ (+4\% vs.\ +16\%). Despite sharing queries across depth, the selected router stays feature-dependent: across five denoising timesteps its mean normalized routing-entropy variation is $2.15\times$ that of the per-layer design (Figure~\ref{fig:attnres_adaptivity}; protocol in Appendix~\ref{app:timestep_adaptivity}).

\noindent\textbf{Block size (Table~\ref{tab:attnres_ablation}d).}
$S \in \{4, 8, 16\}$ yield similar loss point estimates (0.962--0.965), so the short sweep does not identify a preferred span. Memory scales with block count: $S{=}4$ uses +2.9\% while $S{=}16$ uses only +1.6\%. We keep $S{=}8$ as an engineering default because it gives two regular 3:1 attention cycles per aggregation block, preserves more intermediate depth sources than $S{=}16$, and keeps overhead near $S{=}16$.

The selected shared router remains adaptive without an explicit timestep input: its routing entropy varies across denoising stages through the timestep-conditioned source features. Direct perturbation and weight-space analyses provide the complementary explanation. Replacing the learned router offset by its mean or evaluating it at the wrong timestep changes validation loss by at most 0.0002, while the offset has 0.979 mean cross-timestep cosine similarity and a time-varying residual only 13\% as large as its constant component (Appendix~\ref{app:tcond_probe}).

\input{table_latex/attnres_ablation}

\subsection{Efficiency Scaling}
\label{sec:efficiency}
\paragraph{Resolution and router scaling.}
The proxy studies select a shared, timestep-independent router with $S{=}8$. We compare the 25\%-softmax backbone with full softmax under one compiled, no-\attnres{} protocol in which every variant uses its best kernels (fused linear attention, FlashAttention for all softmax layers). This is the standard protocol for all DiT-forward comparisons in the paper.

\vspace{-1em}
\paragraph{Sequence- and model-size scaling.}
Figure~\ref{fig:efficiency_profiling}(a) compares six 480p--1080p shapes and three attention ratios. Under the compiled best-kernels no-\attnres{} protocol, full/25\%-hybrid speedup rises from $1.16\times$ to $2.01\times$. At fixed 720p and 24fps it reaches $3.17\times$ at the 60s tensor shape (1{,}441 frames, Figure~\ref{fig:efficiency_profiling}(b)). Under the same protocol, panel~(c) fixes the 720p/10s shape and jointly grows width and depth from 1.2B to 28.9B. The hybrid is faster at all eight scales, and its absolute saving grows from 128 to 948ms. A controlled decomposition shows the two axes act differently: at fixed width, growing depth from 16 to 60 layers leaves the speedup nearly flat ($1.41\times$), since depth multiplies both variants alike, whereas at fixed depth, growing width shifts it from $1.45\times$ to $1.33\times$ as the width-bound FFN and projection cost outpaces attention (Appendix~\ref{app:model_size_scaling}). The hybrid advantage is thus governed by how much of the compute is sequence-bound, which is exactly the regime long video occupies.

\vspace{-1em}
\paragraph{Model and hardware scaling.}
Matched H100/GB200 profiling shows that the 14B backbone scales across hardware generations. At $736{\times}1280{\times}121$, one hybrid forward decreases from $789.7$ms on H100 to $398.3$ms on GB200 ($1.98\times$). Across the full shape sweep through 125.1K latent tokens, GB200 provides a $1.69$--$2.02\times$ gain (Appendix~\ref{app:hardware_scaling}). The architecture advantage also persists on both devices: replacing the 10-anchor hybrid with full softmax increases Full/Hybrid speedup with 720p duration from $1.40\times$ to $2.73\times$ on H100 and from $1.58\times$ to $3.07\times$ on GB200 (Appendix~\ref{app:architecture_scaling_14b}).

\input{figure_latex/efficiency_profiling}

%% file: table_latex/main_results.tex
\begin{table}[t]
\centering
\caption{\looseness=-1 \textbf{VBench quality comparison.} Green intensity ranks the top three results per metric. Superscripts after method names denote score sources and are defined in Appendix~\ref{app:vbench}. Only the 81-frame \ours{} result is shown here; the appendix also provides the full protocols, extended baselines, and our 121- and 193-frame operating points.}
\label{tab:main_results}
\normalsize
\renewcommand{\arraystretch}{1.08}
\setlength{\tabcolsep}{6pt}
\begin{tabular}{@{}lcccccc@{}}
\toprule
\textbf{Model} & \textbf{Params} & \textbf{Attention} & \textbf{Frames} & \textbf{Total$\uparrow$} & \textbf{Quality$\uparrow$} & \textbf{Semantic$\uparrow$} \\
\midrule
LTX-2.3 (base)$^{\dagger}$ & 22B & Full 3D & 121 & 81.95 & 83.85 & 74.35 \\
Cosmos-3 Nano$^{\dagger}$ & 16B & Full 3D & 93 & 83.13 & 84.29 & 78.50 \\
Lance$^{\dagger}$ & 7.1B & MoT & 81 & 83.18 & 84.09 & \cellcolor{green!8}79.55 \\
Wan 2.1 & 1.3B & Full 3D & 81 & 83.31 & 85.23 & 75.65 \\
HunyuanVideo$^{\star}$ & 13B & Full 3D & 129 & 83.43 & 85.07 & 76.88 \\
Wan 2.1 & 14B & Full 3D & 81 & 83.69 & \cellcolor{green!15}85.59 & 76.11 \\
\prev{} & 2B & Linear & 81 & 84.17 & 84.85 & \cellcolor{green!15}81.46 \\
Wan 2.2$^{\star}$ & A14B & MoE & 81 & \cellcolor{green!8}84.23 & \cellcolor{green!8}85.42 & 79.50 \\
Bernini-R$^{\ddagger}$ & 14B & MoE & 81 & \cellcolor{green!25}\textbf{84.64} & 85.18 & \cellcolor{green!25}\textbf{82.49} \\
\midrule
\textbf{\ours{}} & 5B & Hybrid & 81 & \cellcolor{green!15}84.30 & \cellcolor{green!25}\textbf{85.61} & 79.05 \\
\bottomrule
\end{tabular}
\end{table}

%% file: figure_latex/softmax_ablation.tex
\begin{figure}[!t]
    \centering
    \begin{minipage}{\textwidth}
    \begin{subfigure}[t]{0.325\linewidth}
        \centering
        \includegraphics[width=\linewidth]{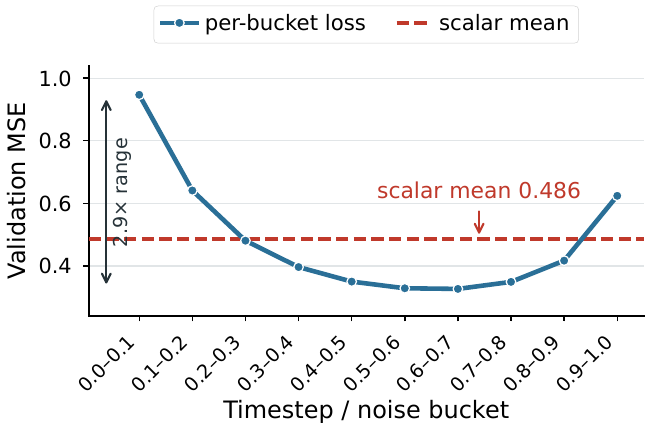}
        \caption{Loss varies with $t$.}
        \label{fig:loss_profile}
    \end{subfigure}
    \hfill
    \begin{subfigure}[t]{0.325\linewidth}
        \centering
        \includegraphics[width=\linewidth]{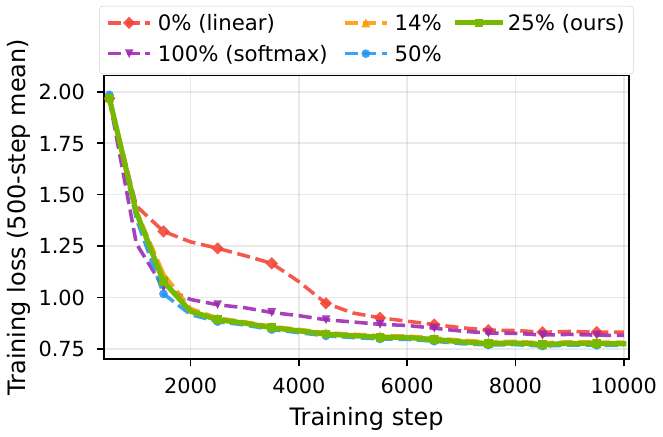}
        \caption{Training loss vs.\ step.}
        \label{fig:loss_curves}
    \end{subfigure}
    \hfill
    \begin{subfigure}[t]{0.325\linewidth}
        \centering
        \includegraphics[width=\linewidth]{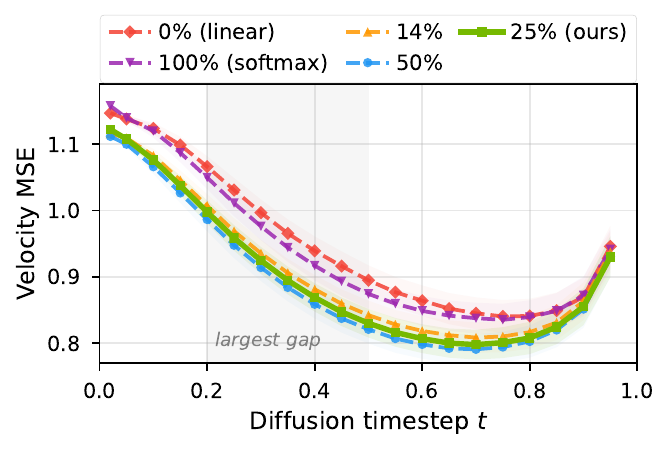}
        \caption{Validation loss vs.\ $t$.}
        \label{fig:per_timestep_loss}
    \end{subfigure}
    \end{minipage}
    \caption{\textbf{Reading loss by timestep, then the softmax-ratio proxy.} (a)~On a production checkpoint, validation loss spans a ${\sim}3\times$ range across the noise axis, so the scalar random-$t$ mean (dashed) hides structure. We therefore read every proxy comparison per timestep rather than as one scalar. (b,c)~Fixed depth-28, width-3,072 architecture-search backbones (256p/81f): (b)~500-step mean training loss and (c)~held-out normalized-latent velocity MSE at 10K (200 videos; pointwise 95\% CIs) for five ratios.}
    \label{fig:softmax_ablation}
    \vspace{-1em}
\end{figure}

%% file: table_latex/attnres_ablation.tex
\begin{table}[H]
\centering
\caption{\textbf{\attnres{} training and design probes.} (a) Late-stage continuation; (b--d) early short-run probes. Wins counts improved noise buckets; green shading marks the selected designs.}
\label{tab:attnres_ablation}
\vspace{-4pt}
\small
\setlength{\tabcolsep}{1.25pt}
    \begin{subtable}[t]{0.297\textwidth}
    \centering
    \caption{Router effect}
    \renewcommand{\arraystretch}{1.20}
    \begin{tabularx}{\linewidth}{@{}>{\raggedright\arraybackslash}Xrr@{}}
    \toprule
    Design & MSE$\downarrow$ & Wins \\
    \midrule
    No \attnres{} & 0.48547 & --- \\
    \rowcolor{green!10}
    + \attnres{} & \textbf{0.48506} & \textbf{17/20} \\
    \bottomrule
    \end{tabularx}
    \end{subtable}%
    \hspace{0.012\textwidth}%
    \begin{subtable}[t]{0.220\textwidth}
    \centering
    \caption{Explicit $t$ input}
    \renewcommand{\arraystretch}{1.20}
    \begin{tabularx}{\linewidth}{@{}>{\raggedright\arraybackslash}Xr@{}}
    \toprule
    Setting & Loss$\downarrow$ \\
    \midrule
    with $t$ & 0.962 \\
    \rowcolor{green!10}
    without $t$ & \textbf{0.920} \\
    \bottomrule
    \end{tabularx}
    \end{subtable}%
    \hspace{0.012\textwidth}%
    \begin{subtable}[t]{0.242\textwidth}
    \centering
    \caption{Query sharing}
    \renewcommand{\arraystretch}{1.20}
    \begin{tabularx}{\linewidth}{@{}>{\raggedright\arraybackslash}Xrr@{}}
    \toprule
    Design & Loss$\downarrow$ & Mem \\
    \midrule
    per-layer & 0.496 & +16\% \\
    \rowcolor{green!10}
    shared & \textbf{0.495} & \textbf{+4\%} \\
    \bottomrule
    \end{tabularx}
    \end{subtable}%
    \hspace{0.012\textwidth}%
    \begin{subtable}[t]{0.182\textwidth}
    \centering
    \caption{Block span ($S$)}
    \renewcommand{\arraystretch}{0.92}
    \begin{tabular}{@{}lrrr@{}}
    \toprule
    $S$ & Loss$\downarrow$ & Lat. & Mem \\
    \midrule
    4 & 0.965 & +3.9\% & +2.9\% \\
    \rowcolor{green!10}
    8 & 0.962 & +3.1\% & +2.1\% \\
    16 & 0.962 & +3.1\% & +1.6\% \\
    \bottomrule
    \end{tabular}
    \end{subtable}
\end{table}

%% file: figure_latex/efficiency_profiling.tex
\begin{figure}[t]
    \centering
    \includegraphics[width=\textwidth]{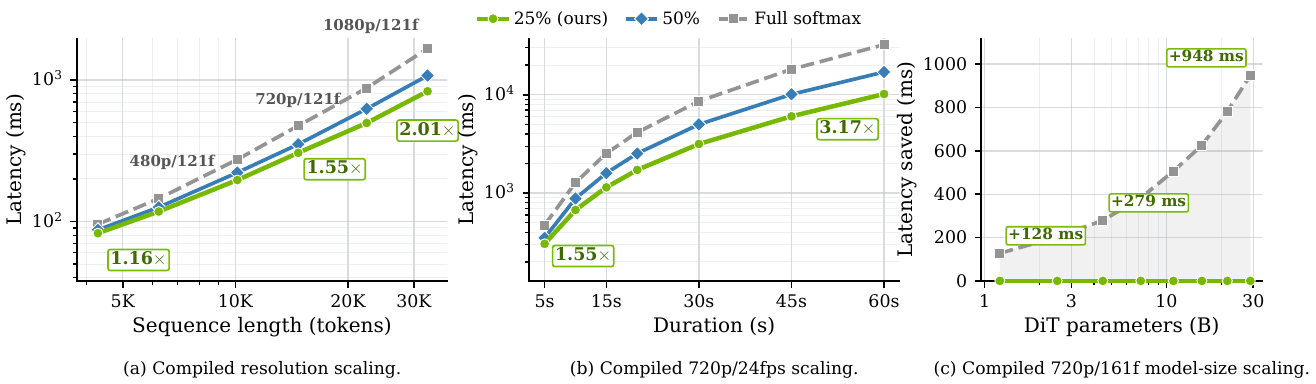}
    \vspace{-1.5em}
    \caption{\looseness=-1 \textbf{Hybrid-attention DiT-forward scaling} (H100, bf16, batch 1; no \attnres{}). (a) Compiled resolution scaling. (b) Compiled 720p/24fps duration scaling; its 121f point is shared with (a). (c) Full-softmax minus 25\%-hybrid latency across model sizes at 720p/161f (19.3K tokens), with the same compiled backend in two run orders. Only (a,b) include 50\% softmax; all are forward-latency profiles.}
    \label{fig:efficiency_profiling}
    \vspace{-1em}
\end{figure}

%% file: sec/06_analysis.tex
\subsection{Mechanistic Analysis}
\label{sec:analysis}
\label{sec:rank_analysis}

\paragraph{State-rank recovery.}
The ratio sweep shows that periodic softmax anchors improve the proxy quality--efficiency tradeoff. We next test whether \attnres{} realizes the complementary goal of carrying richer representations across depth. Using effective rank, the exponential entropy of the linear-state singular spectrum, we toggle depth aggregation on and off on the same production checkpoint under matched prompt and noise seeds. At maximum noise, enabling \attnres{} increases the mean effective rank of the deeper layers by 11.7\%, with non-decreasing rank in 22 of 24 linear layers (Figure~\ref{fig:attnres_rank}). The static router adds a negligible parameter cost (well below 0.001\% of the model).

\vspace{-1em}
\paragraph{Cross-depth reuse.}
On the same architecture-family checkpoint, the router does not collapse to the input or current partial sum: completed-block mass rises with depth and reaches 56\%/50\% for attention/FFN in the deepest blocks (Figure~\ref{fig:attnres_routing}). This reuse is structured rather than diffuse. Within the completed blocks the router favors the most recently finished one, allocating 26.0\%/25.3\% of attention/FFN mass to the nearest completed block versus 15.5\%/13.2\% and 14.5\%/11.1\% to the two earlier blocks, a recency gradient that holds across depth. Removing completed sources at block entries reduces effective rank by 82--91\%, while the same removal at mid-block layers barely changes it (Appendix~\ref{app:rank_ablation}), localizing the reused information to block boundaries where the partial sum has just reset. With the rank recovery above, this evidences cross-depth reuse at the representation level, not only in the routing weights.

\input{figure_latex/attnres_evidence}

\subsection{Online RL Post-Training Results}
\label{sec:post_training_results}

We evaluate the online ReFL stage described in Section~\ref{sec:post_training_online_rl} by tracking its three frozen reward signals throughout training. Across the reported 400-iteration run, the smoothed HPSv3++, DeQA-Score, and UniPercept trajectories all rise together (Figure~\ref{fig:online_rl_reward_curves}), showing that this run improves all three logged components of the joint objective rather than trading one against another. Appendix~\ref{app:online_rl_qualitative} complements these curves with matched-prompt, matched-CFG comparisons between the SFT and RL checkpoints across four examples, providing qualitative context for the changes in visual quality and temporal behavior.

\input{figure_latex/online_rl_reward_curves}

%% file: figure_latex/attnres_evidence.tex
\begin{figure}[htbp]
\centering
\begin{subfigure}[t]{0.305\textwidth}
    \centering
    \includegraphics[width=\linewidth]{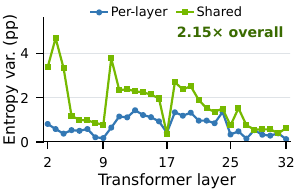}
    \caption{Shared-query adaptivity.}
    \label{fig:attnres_adaptivity}
\end{subfigure}\hfill
\begin{subfigure}[t]{0.416\textwidth}
    \centering
    \includegraphics[width=\linewidth]{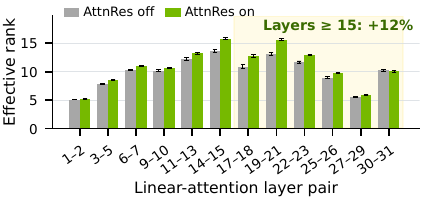}
    \caption{Same-checkpoint rank recovery.}
    \label{fig:attnres_rank}
\end{subfigure}\hfill
\begin{subfigure}[t]{0.260\textwidth}
    \centering
    \includegraphics[width=\linewidth]{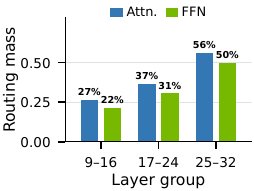}
    \caption{Cross-block reuse.}
    \label{fig:attnres_routing}
\end{subfigure}
\caption{\textbf{Evidence for the selected \attnres{} design.} (a) A shared query remains feature-adaptive across timesteps. (b) Enabling depth aggregation recovers effective rank in deeper linear layers. (c) Learned routing increasingly reuses completed blocks.}
\label{fig:attnres_evidence}
\vspace{-1em}
\end{figure}

%% file: figure_latex/online_rl_reward_curves.tex
\begin{figure}[!t]
    \centering
    \begin{subfigure}[t]{0.325\linewidth}
        \centering
        \includegraphics[width=\linewidth]{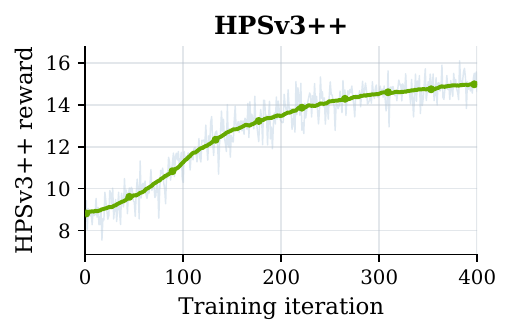}
        \caption{HPSv3++ preference reward.}
        \label{fig:online_rl_hpsv3pp}
    \end{subfigure}
    \hfill
    \begin{subfigure}[t]{0.325\linewidth}
        \centering
        \includegraphics[width=\linewidth]{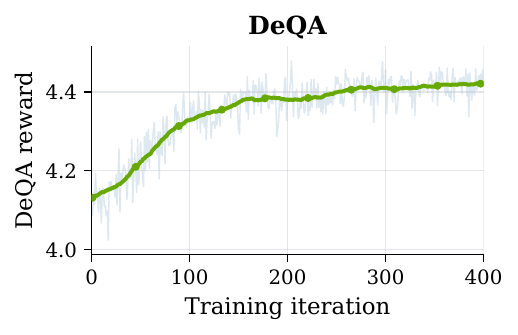}
        \caption{DeQA perceptual quality.}
        \label{fig:online_rl_deqa}
    \end{subfigure}
    \hfill
    \begin{subfigure}[t]{0.325\linewidth}
        \centering
        \includegraphics[width=\linewidth]{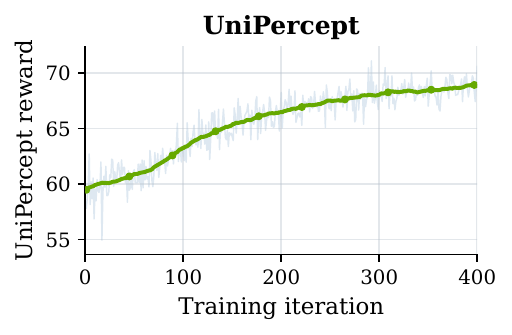}
        \caption{UniPercept mean score.}
        \label{fig:online_rl_unipercept}
    \end{subfigure}
    \caption{\looseness=-1 \textbf{All three reward signals improve steadily during online RL training.} Over 400 ReFL iterations, HPSv3++, DeQA-Score, and UniPercept each exhibit a consistent upward trend; pale curves show per-iteration scores, and green curves show their smoothed trajectories.}
    \label{fig:online_rl_reward_curves}
\end{figure}

%% file: sec/07_applications.tex
\section{Deployment and Applications}
\label{sec:applications}

\paragraph{Kernel-Friendly Deployment with Sol-Engine.}
The backbone stays close to standard attention and FFN primitives: fixed softmax anchors and a conv-free SwiGLU FFN, with no convolution-adapter path. Removing \prev{}'s temporal-convolution FFN avoids a measured $20$--$29\%$ overhead at the larger scale (Appendix~\ref{app:temporal_conv}) and leaves the attention ratio as the main sequence-scaling variable; accordingly, the DiT-forward comparisons in Section~\ref{sec:efficiency} use equally optimized compiled implementations (FlashAttention for softmax and fused linear attention). Because the backbone maps onto these standard, well-optimized kernels, it composes directly with a production deployment stack. We optimize \ours{} through a three-stage Sol-Engine stack~\cite{solengine} on NVIDIA B200 at $736{\times}1280{\times}193$ (720p, 8s): kernel and execution optimization first reduces end-to-end latency from $62.65$ to $30.74$s, residual reuse then evaluates 33 of 50 denoising steps and reaches $20.89$s, and sparse attention on the softmax anchors, which dominate module time at long sequences (Appendix Figure~\ref{fig:runtime_composition}), reaches $17.52$s, for a directly measured $3.58\times$ speedup (Table~\ref{tab:accel_stack}). These are orthogonal deployment improvements rather than modeling contributions: fusion preserves the computation, while caching and sparse attention introduce approximation. Figure~\ref{fig:speedup_compare} shows matched qualitative outputs from the baseline and the complete accelerated pipeline for the same prompts.

\input{figure_latex/speedup_compare}

\input{table_latex/accel_stack}

\paragraph{Low-Precision Inference with QAT.}
We quantize the \ours{} 5B backbone using MXFP4 weights and MXFP8 activations (Table~\ref{tab:qat_quant}). QAT matches the BF16 baseline on VBench Total, while PTQ is slightly lower. On an NVIDIA GB200, one CFG-packed backbone forward for an 81-frame video at $832{\times}480$ resolution decreases from $203.08$ to $191.30$ms ($5.8\%$); static model storage decreases from $8.94$ to $2.87$GB ($67.9\%$), and peak allocated memory from $10.74$ to $4.63$GB ($56.9\%$). The latency gain is modest because the selected GEMMs occupy only $16.2\%$ of BF16 runtime: W4A8 reduces them from $33.15$ to $22.16$ms, and therefore saves only about $11$ms overall. We currently quantize only linear GEMMs; the linear-attention and softmax-attention operators remain in BF16.

\input{table_latex/qat_quant}

\paragraph{Physical AI.}
\ours holds strong potential for Physical AI scenarios, especially efficiency-sensitive tasks in robotics and self-driving. We fine-tune \ours on roughly 5,000 hours of publicly available real-robot and egocentric videos for 100k iterations at a learning rate of $1 \times 10^{-4}$. As shown in Figure~\ref{fig:physical-ai}, \ours produces realistic and physically plausible robot-manipulation videos. It outperforms the similar-sized Cosmos3-Edge~\cite{agarwal2026cosmos} (4B) in the presented comparison and delivers competitive results against much larger models, including Cosmos3-Nano~\cite{agarwal2026cosmos} (16B) and Lingbot-Video~\cite{lingbotvideo} (30B).

\input{figure_latex/physical_ai}

%% file: figure_latex/speedup_compare.tex
\begin{figure}[H]
    \centering
    \includegraphics[width=\textwidth]{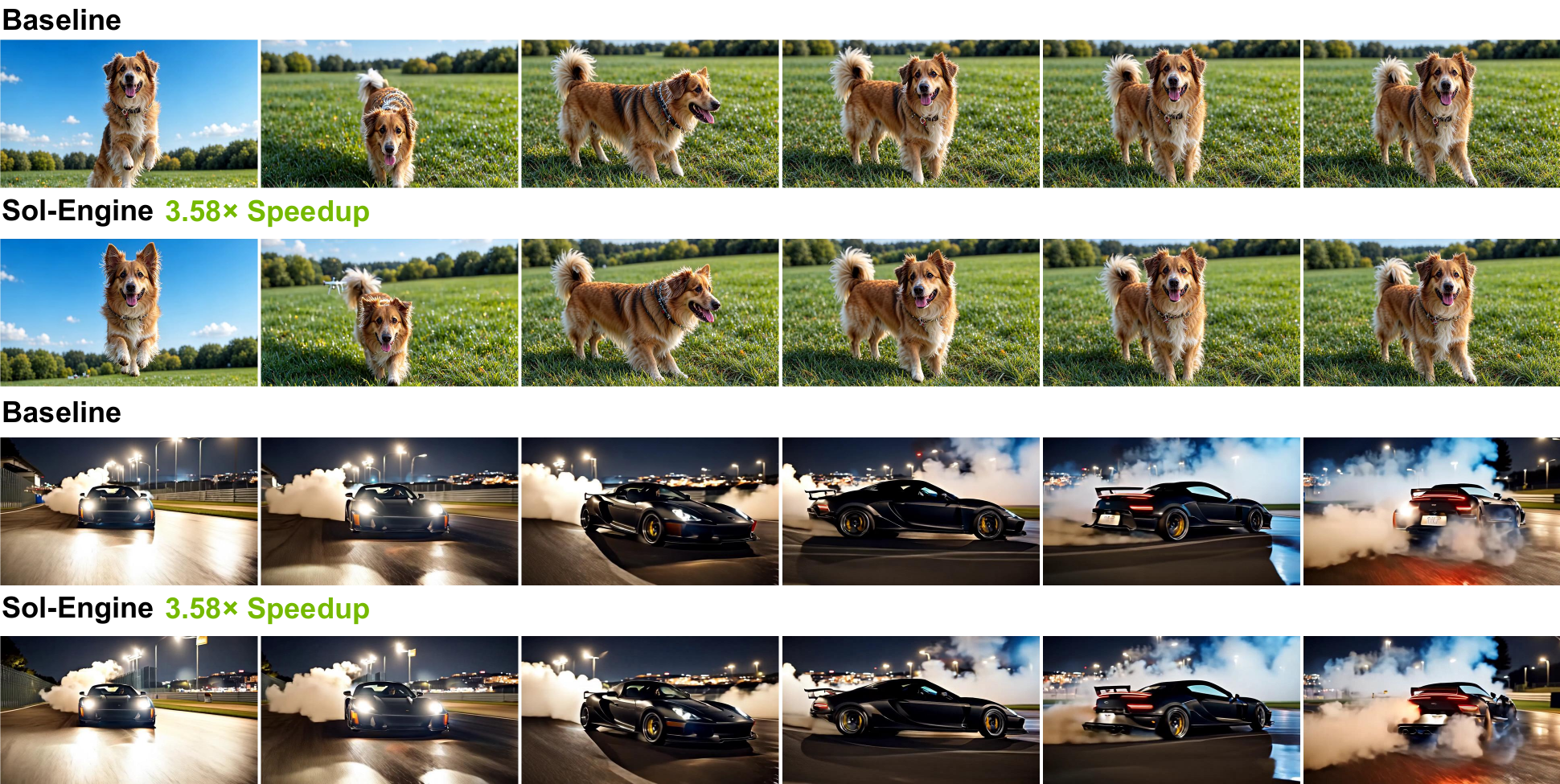}
    \caption{\looseness=-1 \textbf{Qualitative outputs before and after Sol-Engine acceleration.} Two matched prompts compare the baseline with the directly measured $3.58\times$ pipeline; each row samples matched progress through the clip.}
    \label{fig:speedup_compare}
\end{figure}

%% file: table_latex/accel_stack.tex
\begin{table}[htbp]
\centering
\small
\renewcommand{\arraystretch}{1.15} %
\setlength{\tabcolsep}{6pt} %
\caption{\textbf{Full stack acceleration of Sol-Engine} on one B200 and one H100 at 720p resolution and 8s duration. Latency includes denoising and VAE decoding. Speedups are relative to each GPU's own baseline.}
\label{tab:accel_stack}
\begin{tabular}{@{} l *{5}{c} @{}} %
\toprule
\multirow{2}{*}{\textbf{Stage}} & \multirow{2}{*}{\textbf{NFEs}} & \multicolumn{2}{c}{\textbf{B200}} & \multicolumn{2}{c}{\textbf{H100}} \\
\cmidrule(lr){3-4} \cmidrule(lr){5-6} %
 & & \textbf{Latency (s)} & \textbf{Speedup} & \textbf{Latency (s)} & \textbf{Speedup} \\
\midrule
Baseline                    & 50 & 62.65          & 1.00$\times$          & 95.08          & 1.00$\times$ \\
\quad + Kernel Optimization & 50 & 30.74          & 2.04$\times$          & 59.32          & 1.60$\times$ \\
\quad + Diffusion Cache     & 33 & 20.89          & 3.00$\times$          & 40.07          & 2.37$\times$ \\
\quad + Sparse Attention    & 33 & \textbf{17.52} & \textbf{3.58}$\times$ & \textbf{33.43} & \textbf{2.84}$\times$ \\
\bottomrule
\end{tabular}
\end{table}

%% file: table_latex/qat_quant.tex
\begin{center}
\centering
\footnotesize
\captionof{table}{\textbf{QAT quality and low-precision efficiency.} VBench scores (\%) and median systems measurements for one CFG-packed 81-frame backbone forward at $832{\times}480$ on NVIDIA GB200. QAT matches the BF16 baseline, while PTQ is slightly lower; PTQ and QAT share the low-precision inference configuration. Memory uses decimal GB.}
\label{tab:qat_quant}
\setlength{\tabcolsep}{2pt}
\begin{tabular*}{0.9\linewidth}{@{\extracolsep{\fill}}llccccccc@{}}
\toprule
& & \multicolumn{4}{c}{\textbf{VBench}} & \multicolumn{3}{c}{\textbf{GB200 systems}} \\
\cmidrule(lr){3-6}\cmidrule(l){7-9}
\textbf{Method} & \textbf{Precision} & \textbf{Quality} & \textbf{Semantic} & \textbf{Total} & \textbf{$\Delta$} & \textbf{Latency (ms)} & \textbf{Static (GB)} & \textbf{Peak (GB)} \\
\midrule
BF16 (baseline) & BF16 & 83.98 & 80.14 & 83.22 & --- & 203.08 & 8.94 & 10.74 \\
PTQ & MXFP4-W/MXFP8-A & 83.53 & 80.23 & 82.87 & $-0.35$ & \multirow{2}{*}{\textbf{191.30}} & \multirow{2}{*}{\textbf{2.87}} & \multirow{2}{*}{\textbf{4.63}} \\
QAT & MXFP4-W/MXFP8-A & 83.95 & 80.43 & \textbf{83.25} & $+0.03$ & & & \\
\bottomrule
\end{tabular*}
\end{center}

%% file: figure_latex/physical_ai.tex
\begin{figure}[!t]
    \centering
    \includegraphics[width=\textwidth]{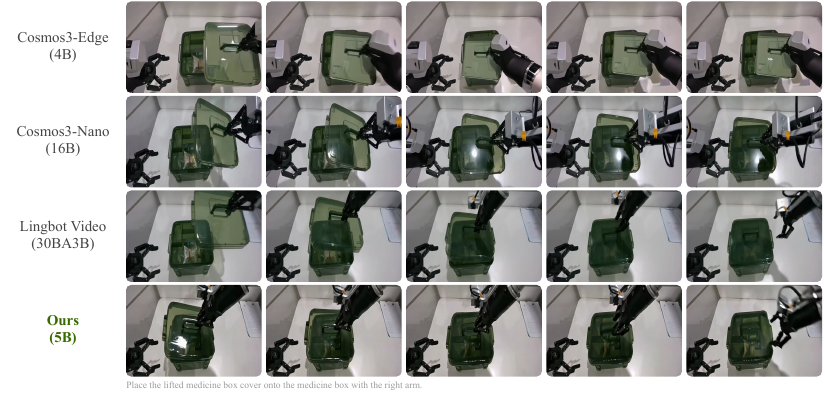}
    \caption{\looseness=-1 \textbf{Qualitative comparison with Physical AI models.} In the presented comparison, \ours{} outperforms the similar-sized Cosmos3-Edge~\cite{agarwal2026cosmos} (4B) and delivers competitive results against much larger models, including Cosmos3-Nano~\cite{agarwal2026cosmos} (16B) and Lingbot-Video~\cite{lingbotvideo} (30B).}
    \label{fig:physical-ai}
\end{figure}

%% file: sec/08_related_work.tex
\section{Related Work}
\label{sec:related}

Most open video generators use diffusion transformers with quadratic softmax token mixing, including Wan~2.1/2.2, HunyuanVideo, CogVideoX, and Open-Sora~\cite{wan2025,hunyuanvideo2024,cogvideox2024,opensora2025}; MAGI-1 and LTX-Video instead reduce effective sequence cost through temporal chunking or stronger latent compression~\cite{magi1,ltxvideo}. Efficient token mixers replace the $N{\times}N$ map with a fixed-size state through linear attention~\cite{katharopoulos2020transformers,yang2024gated,arora2024based} or state-space models~\cite{mamba,videomamba,dim,liu2024vmamba}, although compressed states can restrict representation rank~\cite{zhang2024rala}. In visual generation, DiG applies gated linear attention to image DiTs, \prev{} develops a pure-linear video DiT, and SANA-WM/SANA-Streaming extend efficient backbones to world modeling and streaming editing~\cite{zhu2025dig,sanavideo2025,sanawm2026,sanastreaming2026}. Hybrid language models retain periodic softmax layers: Qwen3-Next and Kimi-Linear use 3:1 layouts, Kimi~K3 adds cross-layer Attention Residuals, and output gating provides a compatible softmax stabilization mechanism~\cite{qwen3next,yang2025kimi,kimik3,kimiteam2026attnres,qiu2025gated}; Attention Surgery studies post-hoc linearization of pretrained video DiTs~\cite{chen2025surgery}. Sparse-VideoGen, Radial Attention, SpargeAttn, PISA, and Native Sparse Attention instead reduce the cost within each remaining softmax operator~\cite{svg2025,radial2025,spargeattn2025,pisa2026,nsa2025}. \ours{} brings hybrid attention with \attnres{} to bidirectional video diffusion, trains it from scratch, re-selects the softmax ratio for video, and adapts depth routing to timestep-conditioned features. Appendix~\ref{app:related_work} presents the complete discussion by research direction.

%% file: sec/09_conclusion.tex
\section{Conclusion}
\label{sec:conclusion}

We introduced \ours{}, a video diffusion transformer built on a mostly-linear hybrid-attention backbone and instantiated at 5B and 14B scales. Instead of quadratic softmax in every layer, it combines 75\% gated linear attention with 25\% periodic softmax anchors, while depth-shared Block \attnres{} propagates the anchors' refreshed representations to later linear layers. Video-specific proxy studies identify 25\% softmax as a practical quality--efficiency knee; the larger production model uses 40 layers at width 4,096 and is trained at 384$\times$B200 scale. With 40-step sampling, the 5B checkpoint reaches VBench Total $84.30$ in $13.2$s at $480{\times}832{\times}81$ on one H100, and under matched best-kernel compilation its DiT forward is $3.2\times$ faster than full softmax at the 720p/60s tensor shape. The hardware-friendly backbone maps cleanly onto fused kernels and composes directly with Sol-Engine's execution, residual-reuse, and sparse-anchor optimizations, yielding a separately measured $3.58\times$ end-to-end speedup in a 50-step B200 deployment. Same-checkpoint analyses show \attnres{} raises deep-layer effective rank by ${\sim}12\%$ and reuses completed-block representations across depth. Although architecture selection relies on short proxy studies and the longest-duration results are tensor-shape profiles, these results show that a mostly-linear hybrid can retain competitive generation quality while scaling far more efficiently to long, high-resolution video.

\medskip
\noindent \textbf{Future work.}
Two directions follow most directly from this design. Because the backbone is $O(N)$-dominated, its efficiency advantage widens with duration (Section~\ref{sec:efficiency}), so extending the curriculum beyond the current 8s horizon toward minute-scale training would turn the profiled long-sequence scaling into genuine long-video generation and stress the anchors' ability to hold global consistency over far longer contexts. Second, our operator is bidirectional, whereas robotics, autonomous driving, and world models require streaming, action-conditioned rollouts: as noted in Section~\ref{sec:prelim}, dropping the delta-rule update makes it a natural initialization for a causal Gated DeltaNet, so pairing a causal delta-rule linear operator in the Kimi-Linear~\cite{yang2025kimi} / Kimi~K3~\cite{kimik3} family with the same hybrid-plus-\attnres{} layout would carry this scratch-trained, kernel-friendly recipe into causal generation and closed-loop world models, complementing the Physical~AI study in Section~\ref{sec:applications}. Further extensions include anchor-specific sparse kernels and calibrated low-precision (NVFP4) formats, few-step distillation to complement the training-free deployment caching, and learned, content-adaptive anchor placement and \attnres{} routing.

\medskip
\noindent \textbf{Acknowledgements.}
We would like to express our sincere gratitude to Yufan Deng (PKU) for invaluable discussions on efficient training. Their constructive feedback and collaboration were instrumental in shaping this work.

%% file: sec/10_references.tex
\bibliographystyle{plainnat}
\bibliography{main}

%% file: sec/11_appendix.tex
\section{Related Work}
\label{app:related_work}

\subsection{Video Diffusion and Efficient Sequence Modeling}

Modern open video generators are largely built on diffusion transformers with full uniform softmax attention. Wan~2.1/2.2, HunyuanVideo, CogVideoX, and Open-Sora scale this design through larger backbones and improved training recipes, with Wan~2.2 adopting a mixture-of-experts denoiser~\cite{wan2025,wan2025v22,hunyuanvideo2024,cogvideox2024,opensora2025}. Their global token mixing is expressive, but its quadratic cost becomes increasingly expensive as resolution and duration grow. Other systems reduce the sequence length instead: MAGI-1 generates video autoregressively in temporal chunks~\cite{magi1}, while LTX-Video relies on latent compression~\cite{ltxvideo}.

Efficient sequence models reduce backbone cost directly. Linear attention replaces the explicit $N{\times}N$ attention map with a fixed-size state~\cite{katharopoulos2020transformers,yang2024gated,arora2024based}, while state-space models use recurrent or selective state updates~\cite{mamba,videomamba,dim,liu2024vmamba}. In vision, DiG applies gated linear attention to image diffusion transformers~\cite{zhu2025dig}; \prev{} develops a pure-linear video DiT with a temporal-convolution FFN~\cite{sanavideo2025}; and SANA-WM and SANA-Streaming extend efficient SANA-family backbones to world modeling and streaming video editing~\cite{sanawm2026,sanastreaming2026}. Because a compressed state can restrict representation rank and direct token interaction~\cite{zhang2024rala}, we keep full-sequence generation but periodically restore exact softmax interaction instead of relying on a pure-linear stack or an additional temporal-convolution path.

\subsection{Hybrid and Sparse Attention}

Hybrid language models combine efficient state-based layers with a smaller number of exact softmax layers. Qwen3-Next and Kimi-Linear use a regular 3:1 linear-to-softmax layout with different linear operators~\cite{qwen3next,yang2025kimi}, and post-attention output gating provides a compatible stabilization mechanism for the softmax branch~\cite{qiu2025gated}. Attention Surgery instead linearizes a pretrained video DiT after training~\cite{chen2025surgery}. We study a bidirectional video diffusion transformer trained from scratch with its hybrid layout fixed throughout training, and re-select the softmax fraction in the video regime rather than assuming that the language-model ratio transfers directly.

Sparse methods provide a complementary way to reduce the remaining softmax cost. Sparse-VideoGen and Radial Attention exploit structure in video attention maps, SpargeAttn and PISA provide training-free sparse or approximate kernels, and Native Sparse Attention learns hardware-aligned sparse patterns for long-context language modeling~\cite{svg2025,radial2025,spargeattn2025,pisa2026,nsa2025}. These methods reduce the work within each softmax layer, whereas our hybrid design reduces how many layers invoke softmax. In our deployment study, sparsity is therefore applied only to the fixed softmax anchors, leaving the linear majority and trained architecture unchanged (Section~\ref{sec:applications}).

\subsection{Cross-Layer Residual Routing}

Standard transformer residual streams pass features only through adjacent layers. Attention Residuals allow a layer to retrieve completed block features from earlier depths, and Kimi~K3 combines this mechanism with hybrid attention at language-model scale~\cite{kimiteam2026attnres,kimik3}. Video diffusion introduces different requirements: features are bidirectional, every block is modulated by a denoising timestep, and the spatial--temporal token set is much larger. We therefore route over completed block summaries and the current partial block, share one query projection across depth, and retain timestep information through the ordinary DiT features rather than a separate router-timestep projection. The routing and rank analyses in this paper test whether the \attnres{} path reuses the representations refreshed by the softmax anchors.

\section{Model and Training Details}
\label{app:config}

\subsection{Model Configuration}

\ours{} has 5B and 14B configurations that share the selected hybrid design. Table~\ref{tab:model_config} records both architectures and their scale-specific training hyperparameters.

\input{table_latex/model_config}

\subsection{Self-Flow Distillation and Dual-Timestep Tokens}
\label{app:selfflow}
Self-Flow~\cite{selfflow2026} is an auxiliary training aid applied only during pre-training and continual training (disabled for SFT). It adds a feature-distillation loss from a shallow student readout toward a deeper (or, in later continual training, EMA) teacher, and a dual-timestep schedule that assigns a small token partition ($R_M{=}0.1$) to a second timestep. All tokens stay in the flow-matching loss, so it changes conditioning rather than reducing backbone compute; the exact layer indices and weights are in Table~\ref{tab:model_config}.

\subsection{Content-aware Flow-Shift}
\label{app:tqd}
TQD~\cite{tqd} uses curation scores to place supervision where each pre-training clip is most informative. With base flow-shift~$1$, clips above the fps-normalized UniMatch threshold of~30 but not the DOVER quality threshold receive a $+1.1$ logit bias toward high noise; clips above the DOVER threshold of~0.91 but not the motion threshold receive a $-1.1$ bias toward low noise; clips satisfying both or neither receive zero bias. The two biased branches have median timesteps of approximately $0.75$ and $0.25$, corresponding to effective shifts of about $3$ and $1/3$, respectively (Figure~\ref{fig:tqd_density}). From continual training onward TQD is disabled and a tail-floored token-aware density with shift $3$--$6$ is used. Shift~3 aligns approximately with the high-motion branch's median, but the distributions and the other TQD branches differ, so this is not a per-sample continuous handoff.

\begin{center}
\centering
\includegraphics[width=0.54\linewidth]{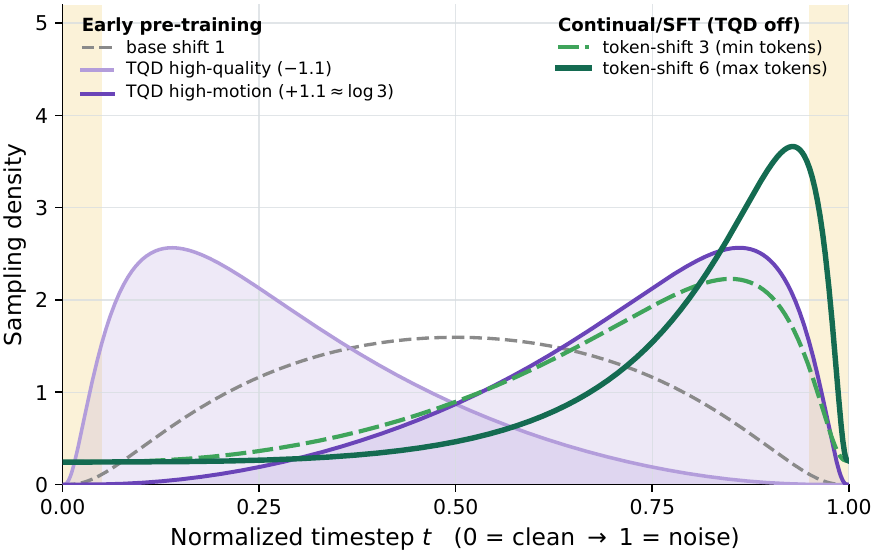}
\captionof{figure}{\textbf{Timestep densities used in successive stages.} Pre-training TQD applies opposite offsets to the mutually exclusive high-quality and high-motion branches. Continual training and SFT disable TQD and use a tail-floored token-count shift spanning $3$--$6$; shift~3 approximately aligns with the high-motion branch's median but is not a continuous per-sample handoff.}
\label{fig:tqd_density}
\end{center}

\subsection{Timestep-Stratified Checkpoint Monitoring}
\label{app:timestep_buckets}

A random-timestep mean can hide improvements in one noise regime behind regressions in another. We therefore divide the 1,000-step noise axis into ten equal buckets and compare four checkpoints from one continuous trajectory. Every checkpoint reuses the same 100 clips, text features, noise tensors, and timestep assignments, covering each integer timestep exactly once. The equal-bucket average is a monitoring statistic, not the training objective.

From the first to last checkpoint, macro MSE falls by $6.4\%$, but the bucket view reveals the useful detail: low-noise MSE improves by as much as $11.4\%$, while the highest-noise bucket regresses by $1.2\%$ (Figure~\ref{fig:timestep_bucket_trajectory}). Stratification also reduces the standard deviation of the change estimate by $2.3\times$ relative to IID timestep sampling. VBench Total rises from 82.68 to 83.29 over these checkpoints; with four observations, this is context rather than a correlation claim.

\input{figure_latex/timestep_bucket_trajectory}

\section{Data Curation and Progressive Selection}
\label{app:data_curation}

We process the corpus with a uniform six-stage pipeline. Its purpose is not only to remove defective clips, but to construct the progressive supervision used by the training curriculum: pre-training favors coverage, continual training raises the quality and motion-consistency floors, and SFT uses the most trusted subset. The same processing logic is used across these pools, with stage-dependent selection thresholds.

\input{table_latex/scoring_metrics}

\paragraph{Decode, segment, and clean.}
We first reject undecodable or undersized clips, normalize metadata, and split long videos into coherent shots with TransNetV2~\cite{transnetv2}. Temporally stable crops remove letterboxing and subtitles when possible; clips dominated by overlays, unstable cuts, severe blur, broken exposure, or temporal artifacts are discarded.

\paragraph{Score complementary properties.}
The remaining clips are evaluated along separate quality, motion, color, and consistency axes rather than collapsed into one score (Table~\ref{tab:scoring_metrics}). DOVER measures visual quality, UniMatch and VMAF-derived features describe motion, and SigLIP features measure visual--text consistency~\cite{wu2023dover,unimatch,vmaf,siglip}. Decisions are aggregated over time, and optical flow is normalized by frame rate so informative low-motion clips are not removed simply for being slow.

\paragraph{Build the curriculum.}
Progressively stricter per-axis thresholds form a broad pre-training pool, a cleaner continual-training pool, and a small trusted SFT subset. Captions follow the same progression, ending in a structured description of subjects, actions, camera motion, scene, lighting, interactions, and temporal evolution. Quantized motion descriptors are appended when available so motion remains explicit to the text conditioner.

\paragraph{Resolution, FPS, and duration schedule.}
The run advances resolution ($480p\to720p$), frame rate, and duration together, with the 720p share increasing in later stages (Table~\ref{tab:training_curriculum}). Frame counts are chosen so each clip lands on an integer number of LTX-VAE latent frames. Because FSDP runs at per-rank batch size one, the sampler buckets each rank into a single (aspect-ratio, frame-count) tier per step, so resolution/duration mixing does not desynchronize ranks. Image batches are interleaved with video at a fixed cadence to inject appearance quality into a motion-biased corpus.

\section{Evaluation and Measurement Protocols}
\label{app:vbench}

\subsection{Sampling and VBench Evaluation}

The reported \ours{} operating points use 40 flow-DPM-Solver steps, motion bucket $30$, seed $0$, and Gemma-2-2B-IT text features~\cite{gemma2}. The 81-frame, 16-fps setting uses classifier-free guidance $6.0$ and flow-shift $6.0$; the 121/193-frame, 24-fps settings use guidance $8.0$ and flow-shift $12.0$. The 81-frame result comes from one late-stage production checkpoint, while the 121/193-frame results share a second late-stage checkpoint. Baselines keep their default settings.

We evaluate the full VBench~\cite{vbench} text-to-video suite with official prompts over all 16 dimensions. Quality averages seven and Semantic nine; Total uses the default $4{:}1$ quality--semantic weighting. Table~\ref{tab:vbench_full} abbreviates the dimensions as SC/BC (subject/background consistency), TF (temporal flickering), MS (motion smoothness), DD (dynamic degree), AQ/IQ (aesthetic/imaging quality), OC (object class), MO (multiple objects), HA (human action), CO (color), SR (spatial relationship), SE (scene), AS/TS (appearance/temporal style), and OV (overall consistency).

\paragraph{Score sources.}
All baseline scores are either official results ($\star$: official 720p leaderboard, $\ddagger$: author-reported) or measured by us with each model's official pipeline and default inference settings ($\dagger$). Unmarked scores use the common $480{\times}832{\times}81$ protocol. Table~\ref{tab:main_results} shows the compact selection with the 81-frame \ours{} result, and Table~\ref{tab:vbench_full} adds the full baseline set and our 121-/193-frame operating points. Wan~2.2 is A14B active (27B total)~\cite{wan2025v22}.

\input{table_latex/vbench_full}

\paragraph{Latency protocol.}
End-to-end latency covers text encoding, denoising, and VAE decoding at batch size one, excluding checkpoint loading and video writing. Unless noted otherwise, we report the mean of three steady-state videos after one full warmup on a single H100. Baselines use their official pipelines and native 720p-class shapes; Wan~2.2-A14B, Bernini-R, and both \ours{} scales use 40 steps in the headline comparison.

At $736{\times}1280{\times}81$, the 5B and 14B configurations take $30.9$s and $69.3$s. At the matched $480{\times}832{\times}81$ shape they take $13.2$s and $29.1$s, versus $421$s for Bernini-R. For context, \prev{}-2B officially reports $36$s at 720p~\cite{sanavideo2025}.

\subsection{Latency and Profiling Protocols}
\label{app:profiling}

The standard architecture profiles use one H100 80GB, bf16, batch size one, eight warmups, and 20 CUDA-event timed forwards (median). Every attention ratio receives its best implementation: fused linear attention, FlashAttention for softmax, and \texttt{torch.compile}. Figures~\ref{fig:efficiency_profiling}(a,b) vary resolution and duration with \attnres{} disabled; panel~(c) follows the same compiled protocol while scaling matched backbones at a fixed 19.3K-token shape. These are DiT-forward measurements and exclude text encoding and VAE decoding.

The module-composition figure is a separate eager diagnostic with the selected shared router enabled. It attributes non-overlapping CUDA-event intervals to top-level modules at 22.1K and 111.3K tokens; it is used only to identify bottlenecks, not as a second speedup protocol or as evidence about learned routing. Results from different protocols are never multiplied unless a table explicitly defines the composition.

\paragraph{14B cross-hardware protocol.}
The H100 and GB200 sweeps use identical software, fixed synthetic inputs, and compiled kernels. The 40-layer, width-4,096 hybrid has 30 fused-linear layers and 10 Flash-SDPA anchors; the full-softmax control sends all 40 layers through the same Flash-SDPA implementation. Width, depth, inputs, and all non-attention settings stay fixed, \attnres{} is disabled in both variants to isolate the attention backbone, and each shape runs in an isolated process. Repeated measurements agree within $0.8\%$ on both devices.

\section{Qualitative Results}
\label{app:qualitative}

\subsection{Additional Qualitative Samples}
\label{app:more_samples}

Figure~\ref{fig:more_samples} shows six additional 720p text-to-video clips ($1280{\times}736$, 193 frames, 8s at 24fps) spanning people, wildlife, urban scenes, and stylized content. Each row unrolls one clip as five frames at $0/2/4/6/8$s, generated with the same 40-step 193-frame sampling recipe as above; the generating prompt is reproduced beneath each row. Four further showcase prompts from the same batches (a violinist in light rain, an octopus changing color, a drone light show, and a desert lightning storm) are held out for the cross-method comparison below (Figure~\ref{fig:t2v_compare}).

\subsection{Qualitative Effect of Online RL}
\label{app:online_rl_qualitative}

Figure~\ref{fig:online_rl_qualitative} complements the training-time reward trajectories in Figure~\ref{fig:online_rl_reward_curves} with four matched-prompt, matched-CFG comparisons between the SFT and RL checkpoints. Five evenly spaced frames expose changes throughout each eight-second clip; these prompt-level examples provide qualitative context and are not an aggregate evaluation.

\subsection{Qualitative Comparison with Other Methods}
\label{app:method_compare}

Figure~\ref{fig:t2v_compare} compares \ours{} with Wan~2.2-A14B, Bernini-R~14B, Cosmos-3 Nano, and LTX-2.3 on four prompts chosen to expose temporal behavior: sustained action, color change, shape transformation, and repeated events. Each method uses its official pipeline, native 720p-class shape, and step count on one H100 (seed~42); \ours{} uses 40 steps and 193 frames. Because durations differ, we show each output at normalized start, midpoint, and end rather than at absolute timestamps. These are qualitative examples, not an aggregate ranking.

Figures~\ref{fig:ti2v_compare_a} and~\ref{fig:ti2v_compare_b} extend the comparison to image-conditioned generation. Every method receives the same first frame and caption from VBench-I2V, and each row shows normalized progress at $0/25/50/100\%$. \ours{} generates 121 frames with its image-conditioned checkpoint; LTX-2.3, Wan~2.2 TI2V-5B, and Cosmos-3 use their official first-frame conditioning paths. Bernini-R is omitted because its public pipeline has no corresponding mode. All methods run at native 720p-class shapes with seed~42 on one H100. The examples illustrate differences in motion and composition retention, but do not substitute for a quantitative TI2V benchmark.

\section{\attnres{} Routing and Representation Analysis}
\label{app:entropy_metrics}

This section explains what the selected router learns and why its final form shares one query across depth without an explicit timestep input. Unless noted otherwise, the analyses use 5B architecture-family checkpoints; the continued-training comparison below uses the mature 4.5B production ancestor.

\subsection{Late-Stage Continued-Training Probe}
\label{app:attnres_continuation}

To test the router after useful video representations have formed, we continue a block-level \attnres{} run from the same mature 4.5B hybrid ancestor as the no-router trajectory. Near-matched checkpoints are evaluated on the same 100 held-out clips and 20 noise levels. Their mean MSE is effectively tied (0.4851 with \attnres{} versus 0.4855 without it), although 17 of 20 buckets slightly favor \attnres{}. We use this probe only to rule out degradation; the evidence for cross-depth reuse comes from the routing and representation interventions below, not from this narrow loss difference.

\subsection{Routing Selectivity Metric}
\label{app:norm_entropy}

For each layer $l$, \attnres{} computes aggregation weights $\alpha_l = \mathrm{softmax}(q_l \cdot K)$ over $N_l$ depth sources (initial embedding $b_0$, completed block summaries $b_1,b_2,\dots$, and the current partial sum $p_l$). We measure selectivity via normalized Shannon entropy:
\begin{equation}
    \hat{H}_l = \frac{H(\alpha_l)}{\log N_l} = \frac{-\sum_{j=1}^{N_l} \alpha_{l,j} \log \alpha_{l,j}}{\log N_l}
\end{equation}
For layer $l$, the attention branch has $N_l^{\mathrm{attn}}=\lceil l/S\rceil+\mathbf{1}[(l-1)\bmod S>0]$ sources, while the FFN branch has $N_l^{\mathrm{ffn}}=\lceil l/S\rceil+1$. We omit the first attention routing site ($N=1$) from normalized-entropy summaries. The normalization otherwise makes source sets of different sizes comparable: $\hat{H}_l=1$ is uniform and $\hat{H}_l\to0$ is selective.

\subsection{Timestep Adaptivity}
\label{app:timestep_adaptivity}

To test timestep-dependent depth aggregation in \attnres{}, we evaluate ten held-out batches at five timesteps with real text conditioning. After averaging per timestep, we compute the variation at each routing site $s$:
\begin{equation}
    \mathrm{Adaptivity}_s = \sigma_t[\hat{H}_s].
\end{equation}
Higher values indicate greater change across denoising stages. Despite using one query projection across depth, the selected shared router shows $2.15\times$ more entropy variation than the per-layer alternative (Figure~\ref{fig:attnres_adaptivity}). This variation comes from timestep-conditioned features in the ordinary DiT pathway; it needs no router-timestep offset.

\subsection{Timestep Conditioning Probe}
\label{app:tcond_probe}

To determine whether the explicit router timestep projection learns timestep-specific information or merely a constant bias, we combine a behavioural probe with a direct weight-space analysis of its learned offset $\phi_\tau(t)$.

\noindent\textbf{Behavioural probe.} Let $q_0$ be the base routing query. We evaluate the ablation checkpoint under three conditions: \textbf{Normal}, $q=q_0+\phi_\tau(t)$; \textbf{Mean}, $q=q_0+\bar{\phi}_\tau$, where $\bar{\phi}_\tau$ averages 21 uniformly spaced timesteps; and \textbf{Shuffle}, $q=q_0+\phi_\tau(1-t)$ for \attnres{} while AdaLN still receives the correct $t$. Their respective aggregate validation-loss point estimates are $0.4905$, $0.4906$, and $0.4907$, a maximum difference of $0.0002$ in this probe.

\noindent\textbf{Weight-space probe (Figure~\ref{fig:tquery_constancy}).} Across 21 timesteps, the learned offset remains nearly constant (mean cosine $0.979$), and its varying component is only $13\%$ of the constant component's norm. Together with the negligible behavioural change above, this motivates removing the explicit router-timestep projection while retaining timestep modulation inside every DiT block.

\input{figure_latex/timestep_query_constancy}

\subsection{Depth-Routing Patterns}
\label{sec:borrows}

Figures~\ref{fig:attnres_routing} and~\ref{fig:routing_source_diagnostics} average routing over ten validation clips. Completed summaries receive roughly half of the routing mass in layers 25--32 (56\% for attention and 50\% for FFN), and the detailed view favors the most recently completed block. The difference at block entries follows operation order: attention routing occurs before the new partial sum exists, whereas FFN routing occurs afterward. Because each summary contains attention, cross-attention, and FFN updates, this evidence supports cross-depth feature reuse without assigning it to a single operator.

\input{figure_latex/routing_source_diagnostics}

\subsection{Completed-Block Routing Ablation}
\label{app:rank_ablation}
We zero the routing weights assigned to completed block sums, renormalize the remaining sources, and measure the effective rank of $h_l$ over the same input at $t\in\{0.3,0.5,0.7\}$ (Figure~\ref{fig:rank_ablation}). At block-entry layers ($9,17,25$), the current partial sum has reset, so removal leaves the input embedding and reduces rank by $82$--$91\%$. The effect becomes small after the partial sum has rebuilt within a block. This localizes completed-block reuse to the boundary where it is needed most, without attributing the recovered rank to one operator inside the block.

\subsection{Same-Checkpoint Rank Probe}
\label{app:rank_probe}

For the gated, RoPE-rotated linear state $S{=}\sum_n v_n(\beta_n k_n^r)^\top$, we define
\begin{equation}
r_{\mathrm{eff}}(S)=\exp\!\left(-\sum_i p_i\log p_i\right),
\qquad p_i=\frac{\sigma_i(S)}{\sum_j\sigma_j(S)}.
\end{equation}
In implementation, we first average corresponding singular values across attention heads, then normalize the averaged spectrum and compute $r_{\mathrm{eff}}$. We toggle depth aggregation on and off on the same production checkpoint and prompt, pairing noise seeds 0--3 between the two conditions. Figure~\ref{fig:attnres_rank} reports the mean and standard deviation across seeds at each linear layer, and the deep-layer summary averages layers ${\ge}15$. No weights or activations other than the \attnres{} path are changed, isolating a representation-level effect of the trained router. For context (Figure~\ref{fig:rank_pure_linear}), we place this result beside a public 2B pure-linear model and a separately trained no-\attnres{} model. These differ in width, training horizon, and checkpoint, so the comparison is contextual rather than an \attnres{} effect estimate.

\input{figure_latex/routing_rank_diagnostics}

\section{Efficiency and Deployment Analysis}
\label{app:efficiency}

This section complements the main efficiency figure with controlled analyses that explain the observed scaling and with direct deployment measurements. Appendix~\ref{app:profiling} defines the protocols; results from different subsections are not composed unless a table explicitly reports the combined measurement.

\subsection{Model-Size Scaling at Fixed Sequence Length}
\label{app:model_size_scaling}

Figure~\ref{fig:efficiency_profiling}(c) fixes a 720p/10s sequence while jointly increasing width and depth. Absolute time saved by the 25\% hybrid grows from 128 to 948ms, while Full/Hybrid speedup narrows from $1.88\times$ to $1.45\times$. A controlled decomposition explains the trend: depth scales both layouts nearly equally, whereas width shifts more compute into shared FFN/projections. These matched forward profiles isolate systems scaling.

\subsection{14B Cross-Hardware Scaling}
\label{app:hardware_scaling}

Figure~\ref{fig:hardware_scaling_14b} compares the same 40-layer, width-4,096 14B hybrid implementation on H100 and GB200. Across the measured resolution and duration range, GB200 reduces forward latency by $1.69$--$2.02\times$ (median $1.91\times$), including the 125K-token endpoint. Peak allocated memory rises from approximately 27 to 44GiB on both devices across this sweep. This establishes that the long-sequence implementation transfers across hardware generations rather than relying on one device-specific operating point.

\input{figure_latex/hardware_scaling_14b}

\subsection{14B Hybrid versus Full-Softmax Scaling}
\label{app:architecture_scaling_14b}

Figure~\ref{fig:architecture_scaling_14b} compares parameter-matched hybrid and full-softmax versions of the 40-layer 14B backbone. As duration grows from 5 to 45s, Full/Hybrid speedup rises from $1.40\times$ to $2.73\times$ on H100 and from $1.58\times$ to $3.07\times$ on GB200. The result confirms that the mostly-linear advantage persists at the larger architecture scale and strengthens with sequence length.

\input{figure_latex/architecture_scaling_14b}

\subsection{Temporal-Convolution FFN Ablation}
\label{app:temporal_conv}

\prev{} uses an additional temporal-convolution path inside its FFN, whereas \ours{} relies on attention for spatiotemporal mixing and keeps a standard SwiGLU FFN. To measure the cost of that choice, we hold depth and width fixed, change only the FFN path, and sweep 720p duration at two model scales (Table~\ref{tab:temporal_conv_scaling}). The variants are architecture-matched rather than parameter-matched because the temporal convolution adds weights. Its overhead grows with duration and reaches $20$--$29\%$ at the larger scale. The large-scale 60s temporal-conv timing is an isolated outlier and is excluded together with its paired baseline. Removing the temporal path simplifies the backbone and preserves the scaling benefit of the selected attention ratio.

\subsection{Module-Level Composition}
\label{app:module_composition}

Figure~\ref{fig:runtime_composition} reports mutually exclusive top-level module shares from an eager shape diagnostic (modules are non-overlapping, so the shares are additive). From 22.1K to 111.3K tokens, the eight softmax anchors grow from 23.0\% to 54.4\% of forward time, while linear self-attention falls from 26.0\% to 15.7\%, which motivates anchor-specific kernels for long shapes.

\input{figure_latex/temporal_conv_runtime}

\subsection{Compile-Friendly \attnres{}}
\label{app:compile_attnres}
The original inference path mutates a source buffer and reduces a growing prefix, which is difficult for \texttt{torch.compile} to fuse. We instead expose the at-most-five sources as a static list and accumulate their weighted sum in fp32 without stacking all sources. Relative to the buffer path, full-forward drift is negligible in fp32 and $0.68\%$ in bf16. The rewrite improves compiled/eager speedup from $1.47\times$ to $2.11\times$ at 8s and from $1.21\times$ to $1.49\times$ at 60s. These numbers compare router implementations, not complete model deployments.

\input{figure_latex/appendix_samples}
\input{figure_latex/online_rl_qualitative}
\input{figure_latex/method_compare_t2v}
\input{figure_latex/ti2v_compare_a}
\input{figure_latex/ti2v_compare_b}

%% file: table_latex/model_config.tex
\begin{table}[H]
\centering
\caption{\textbf{Full 5B and 14B configurations of \ours{}.} Both use the same hybrid-attention design; the 14B settings are verified against the saved Poly training configuration and runtime log.}
\label{tab:model_config}
\small
\setlength{\tabcolsep}{6pt}
\begin{tabularx}{\textwidth}{@{}l>{\centering\arraybackslash}X>{\centering\arraybackslash}X@{}}
\toprule
\textbf{Setting} & \textbf{5B} & \textbf{14B} \\
\midrule
\multicolumn{3}{@{}l}{\textit{Backbone architecture}} \\
Depth & 32 & 40 \\
Hidden dimension & 2,560 & 4,096 \\
Linear attention & 20 heads ($d_h{=}128$) & 32 heads ($d_h{=}128$) \\
Softmax attention & 10 heads ($d_h{=}256$) & 16 heads ($d_h{=}256$) \\
Attention operators & \multicolumn{2}{c}{Gated bidirectional linear / gated softmax; self- and cross-attention QK normalization} \\
Softmax anchors & 8 layers; 25\% (3:1) & 10 layers; 25\% (3:1) \\
\attnres{} & $S{=}8$, shared projection, no time conditioning & $S{=}8$, shared projection, no time conditioning \\
FFN & SwiGLU, ratio 4.0 & SwiGLU, ratio 4.0 \\
Position encoding & \multicolumn{2}{c}{Wan RoPE; separate linear/softmax head dimensions} \\
Patch size & $(1,1,1)$ & $(1,1,1)$ \\
Parameters & ${\sim}4.5$B (5B class) & 14.247B (14B class) \\
\midrule
\multicolumn{3}{@{}l}{\textit{Training configuration}} \\
Resolution / frames & 480p--720p; 81/121/193 & 480p; 81/121 \\
VAE & \multicolumn{2}{c}{LTX-VAE~2.3, 128 channels, causal encode} \\
VAE stride & \multicolumn{2}{c}{$8{\times}32{\times}32$ (temporal $\times$ spatial $\times$ spatial)} \\
Text encoder & \multicolumn{2}{c}{Gemma-2-2B-IT, 300 tokens; normalized features, scale 0.01} \\
Objective & Flow matching; stage-specific TQD / Self-Flow & Flow matching + Self-Flow \\
Self-Flow & student/teacher 9/25, weight 0.8, $R_M{=}0.1$ & student/teacher 14/34, weight 0.8, $R_M{=}0.1$ \\
Noise distribution & shift 1 then token-aware 3--6, $\sigma{=}0.95$ & tail-floored logit-normal, shift 3, $\sigma{=}0.95$ \\
Optimizer & AdamW, lr $10^{-4}$ / $5{\times}10^{-5}$ & AdamW, lr $10^{-4}$, $\beta{=}(0.9,0.999)$, wd 0 \\
Schedule / clipping & constant + 500-step warmup; clip 0.1 & constant + 500 configured warmup steps; clip 0.1 \\
Per-rank batch & 1 video & 1 video / 4 images; no accumulation \\
Precision / sharding & bf16, FSDP & bf16, FSDP + activation checkpointing \\
Weight EMA & enabled & 0.9999 \\
Training hardware & 64$\times$H100 & 384$\times$B200 (Poly) \\
\bottomrule
\end{tabularx}
\end{table}

%% file: figure_latex/timestep_bucket_trajectory.tex
\begin{figure}[htbp]
\centering
\includegraphics[width=0.98\textwidth]{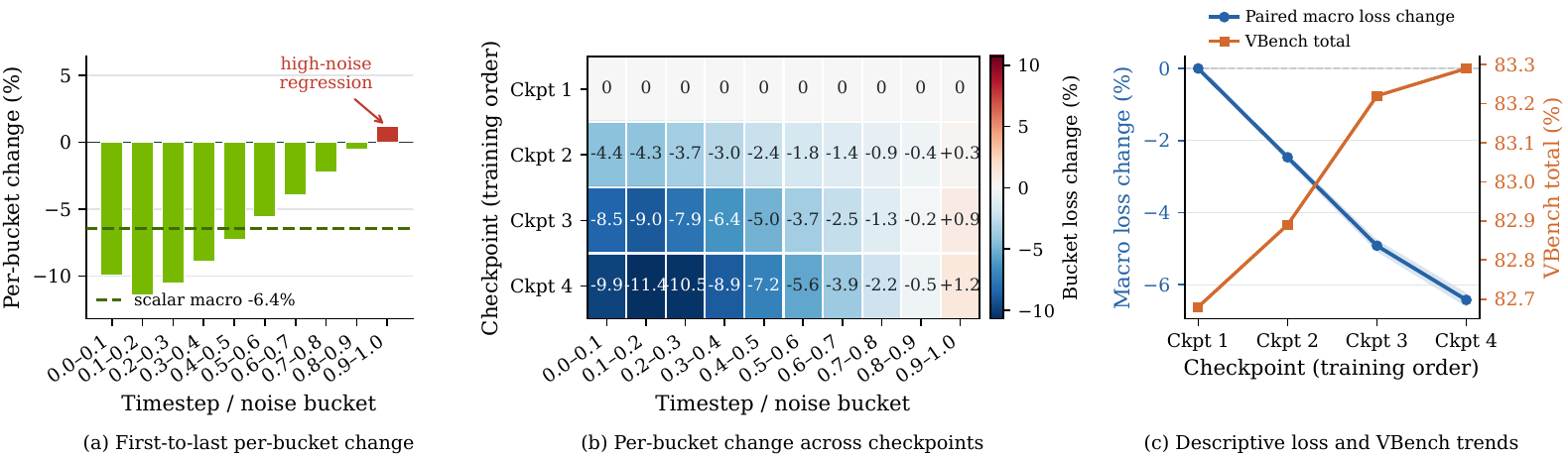}
\caption{\textbf{Timestep-stratified checkpoint monitoring.} Four matched checkpoints from one run. (a,b) Bucketwise loss changes expose a large low-noise improvement and a small high-noise regression that the scalar mean blurs together. (c) Paired macro change with 95\% intervals and the descriptive VBench trend.}
\label{fig:timestep_bucket_trajectory}
\end{figure}

%% file: table_latex/scoring_metrics.tex
\begin{center}
\centering
\small
\captionof{table}{\looseness=-1 \textbf{Signals used by the in-house quality/motion funnel.} Selection thresholds tighten progressively across the pre-training, continual, and SFT stages.}
\label{tab:scoring_metrics}
\begin{tabular}{@{}lll@{}}
\toprule
\textbf{Signal family} & \textbf{Examples} & \textbf{Role} \\
\midrule
Visual quality & DOVER, blur/exposure statistics & Fidelity and aesthetics \\
Motion & optical flow, VMAF motion & Richness and consistency \\
Color & saturation and luminance statistics & Natural appearance \\
Consistency/alignment & image--video, vision--language similarity & Conditioning reliability \\
Shot integrity & cut and artifact detectors & Temporal continuity \\
\bottomrule
\end{tabular}
\end{center}

%% file: table_latex/vbench_full.tex
\begin{center}
\centering
\captionof{table}{\textbf{Complete VBench comparison across all 16 dimensions} (\%). Green intensity ranks the top three results per column (best in bold); dashes mark models with only aggregate scores reported. Appendix~\ref{app:vbench} defines the score-source markers and protocols.}
\label{tab:vbench_full}
\setlength{\tabcolsep}{2.6pt}
\resizebox{\textwidth}{!}{%
\begin{tabular}{@{}l*{19}{c}@{}}
\toprule
\textbf{Model} & \multicolumn{3}{c}{\textbf{Aggregates}} & \multicolumn{7}{c}{\textbf{Quality dimensions}} & \multicolumn{9}{c}{\textbf{Semantic dimensions}} \\
\cmidrule(lr){2-4}\cmidrule(lr){5-11}\cmidrule(l){12-20}
 & Total$\uparrow$ & Quality$\uparrow$ & Semantic$\uparrow$ & SC & BC & TF & MS & DD & AQ & IQ & OC & MO & HA & CO & SR & SE & AS & TS & OV \\
\midrule
LTX-2.3 (base)$^{\dagger}$ & 81.95 & 83.85 & 74.35 & 93.10 & 95.77 & 98.92 & 98.71 & 71.39 & 62.13 & 67.51 & 88.89 & 60.26 & 95.80 & 81.31 & 68.59 & 51.38 & 21.11 & 24.31 & 25.91 \\
CogVideoX 1.5 & 82.17 & 82.78 & 79.76 & 96.87 & 97.35 & 98.88 & 98.31 & 50.93 & 62.79 & 65.02 & 87.47 & 69.65 & \cellcolor{green!15}97.20 & 87.55 & \cellcolor{green!25}\textbf{80.25} & 52.91 & \cellcolor{green!25}\textbf{24.89} & \cellcolor{green!15}25.19 & \cellcolor{green!15}27.30 \\
Cosmos-3 Nano$^{\dagger}$ & 83.13 & 84.29 & 78.50 & 95.05 & 97.26 & 99.08 & \cellcolor{green!25}\textbf{99.06} & 60.28 & 62.71 & \cellcolor{green!8}69.37 & 89.29 & \cellcolor{green!8}78.03 & \cellcolor{green!8}96.60 & \cellcolor{green!8}89.20 & 72.72 & 54.56 & 20.98 & 24.49 & 26.78 \\
Lance$^{\dagger}$ & 83.18 & 84.09 & 79.55 & 94.68 & 94.95 & 98.34 & \cellcolor{green!8}98.92 & 71.11 & 63.71 & 67.71 & \cellcolor{green!8}94.18 & 76.69 & \cellcolor{green!25}\textbf{97.60} & 81.33 & \cellcolor{green!15}79.56 & 52.83 & \cellcolor{green!15}23.12 & \cellcolor{green!8}24.58 & 26.89 \\
Wan 2.1 (1.3B) & 83.31 & 85.23 & 75.65 & \cellcolor{green!8}97.56 & \cellcolor{green!8}97.93 & \cellcolor{green!25}\textbf{99.55} & 98.52 & 65.19 & 65.46 & 67.01 & 88.81 & 74.83 & 94.00 & \cellcolor{green!8}89.20 & 73.04 & 41.96 & 21.81 & 23.13 & 25.50 \\
HunyuanVideo$^{\star}$ & 83.43 & 85.07 & 76.88 & 97.22 & 97.60 & 99.39 & \cellcolor{green!15}99.05 & 71.94 & 60.28 & 67.24 & 83.48 & 66.71 & 94.40 & \cellcolor{green!25}\textbf{89.79} & 72.13 & 54.46 & 22.21 & 24.52 & 26.95 \\
Wan 2.1 (14B) & 83.69 & 85.59 & 76.11 & 97.52 & \cellcolor{green!25}\textbf{98.09} & \cellcolor{green!15}99.46 & 98.30 & 65.46 & 66.07 & \cellcolor{green!15}69.43 & 86.28 & 69.58 & 95.40 & 88.59 & 75.39 & 45.75 & 22.64 & 23.19 & 25.91 \\
\prev{} & 84.17 & 84.85 & \cellcolor{green!15}81.46 & 97.13 & 97.71 & 98.37 & 98.20 & 69.17 & \cellcolor{green!15}68.27 & 64.82 & \cellcolor{green!25}\textbf{95.39} & \cellcolor{green!25}\textbf{85.05} & 95.40 & \cellcolor{green!15}89.43 & 78.00 & \cellcolor{green!8}57.83 & 22.43 & 24.31 & \cellcolor{green!8}27.04 \\
Wan 2.2 (A14B)$^{\star}$ & 84.23 & 85.42 & 79.50 & 97.29 & 97.39 & 99.22 & 98.16 & 61.02 & 67.22 & \cellcolor{green!25}\textbf{71.75} & 94.06 & \cellcolor{green!15}82.10 & 96.40 & 87.43 & \cellcolor{green!8}78.39 & 56.80 & 20.39 & 23.64 & 26.12 \\
Open-Sora 2.0 & 84.34 & 85.40 & \cellcolor{green!8}80.12 & \cellcolor{green!15}97.71 & \cellcolor{green!15}98.00 & \cellcolor{green!8}99.40 & 98.69 & 71.39 & 64.39 & 65.66 & \cellcolor{green!15}94.50 & 77.72 & 95.40 & 85.98 & 76.18 & 52.71 & \cellcolor{green!8}22.98 & \cellcolor{green!25}\textbf{25.91} & \cellcolor{green!25}\textbf{27.50} \\
Bernini-R$^{\ddagger}$ & \cellcolor{green!15}84.64 & 85.18 & \cellcolor{green!25}\textbf{82.49} & -- & -- & -- & -- & -- & -- & -- & -- & -- & -- & -- & -- & -- & -- & -- & -- \\
\midrule
\textbf{\ours{} (81f)} & 84.30 & \cellcolor{green!8}85.61 & 79.05 & 97.34 & 97.33 & 97.84 & 98.31 & \cellcolor{green!8}78.89 & 67.99 & 66.51 & 92.80 & 76.75 & 94.40 & 88.05 & 71.85 & \cellcolor{green!25}\textbf{59.88} & 21.35 & 24.05 & 26.93 \\
\textbf{\ours{} (121f)} & \cellcolor{green!25}\textbf{85.29} & \cellcolor{green!25}\textbf{86.98} & 78.51 & 97.44 & 97.09 & 97.58 & 98.40 & \cellcolor{green!25}\textbf{92.50} & \cellcolor{green!25}\textbf{68.85} & 68.39 & 92.39 & 74.38 & 95.00 & 89.17 & 69.34 & \cellcolor{green!15}58.81 & 21.26 & 24.46 & 26.64 \\
\textbf{\ours{} (193f)} & \cellcolor{green!8}84.48 & \cellcolor{green!15}86.51 & 76.37 & \cellcolor{green!25}\textbf{98.00} & 96.94 & 97.94 & 98.52 & \cellcolor{green!15}88.33 & \cellcolor{green!8}68.10 & 66.31 & 91.76 & 72.39 & 92.60 & 88.21 & 65.32 & 54.68 & 21.27 & 23.48 & 26.08 \\
\bottomrule
\end{tabular}}
\end{center}

%% file: figure_latex/timestep_query_constancy.tex
\begin{figure}[htbp]
\centering
\includegraphics[width=\textwidth]{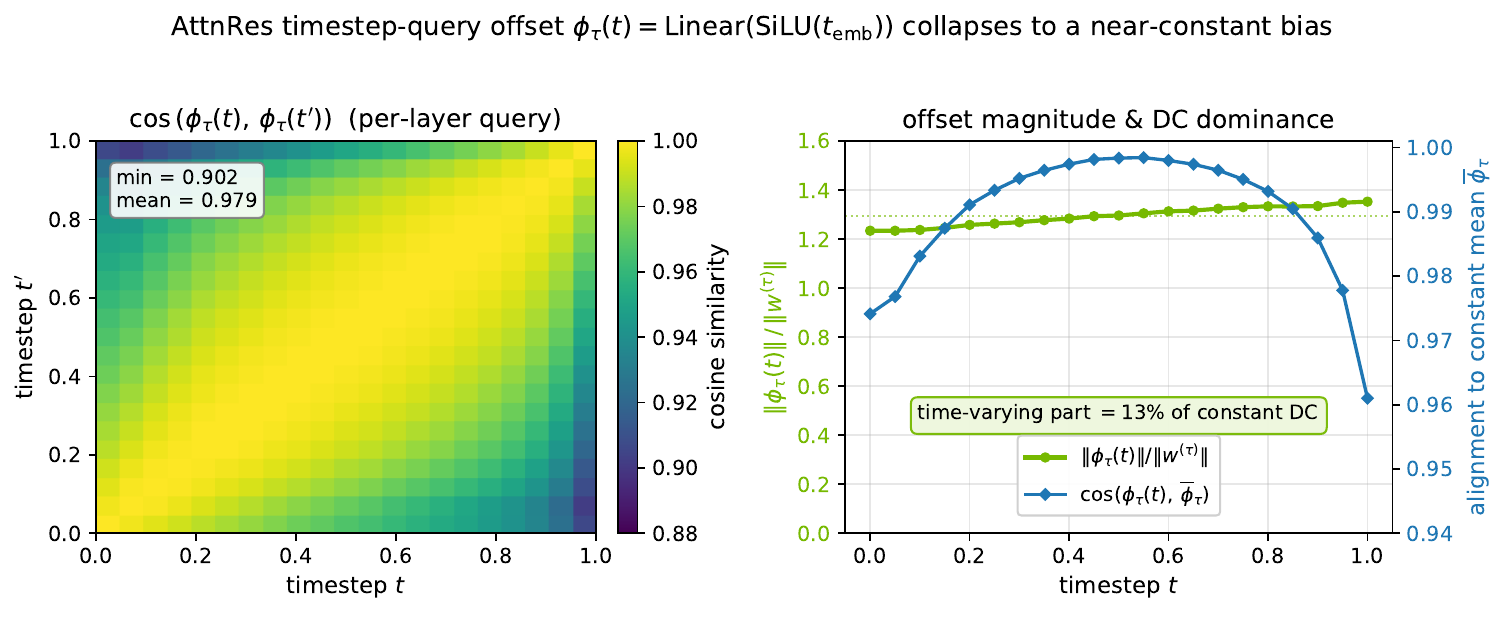}
\caption{\textbf{The explicit router offset is nearly timestep-invariant.} Across 21 timesteps, $\phi_\tau(t)$ remains strongly aligned with its mean and its time-varying component is small.}
\label{fig:tquery_constancy}
\end{figure}

%% file: figure_latex/routing_source_diagnostics.tex
\begin{figure}[tbp]
\centering
\begin{subfigure}[t]{0.73\linewidth}
    \centering
    \includegraphics[width=\linewidth]{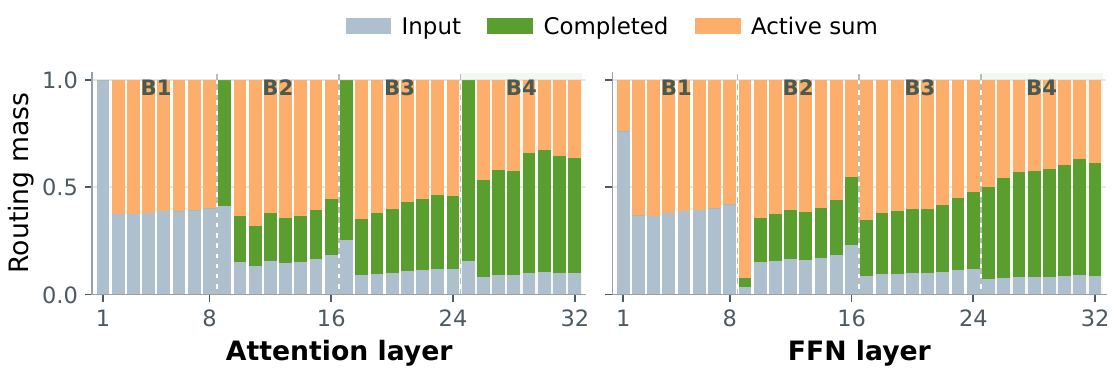}
    \caption{Layerwise routing; B1--B4 mark the 8-layer blocks.}
    \label{fig:routing_source_composition}
\end{subfigure}\hfill
\begin{subfigure}[t]{0.25\linewidth}
    \centering
    \includegraphics[width=\linewidth]{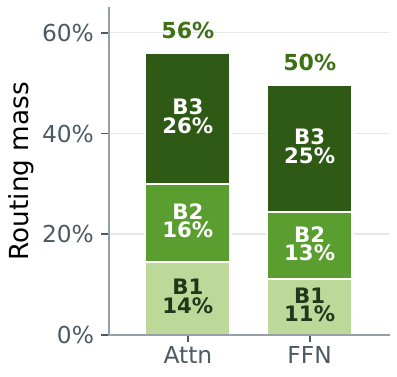}
    \caption{Completed-source detail in layers 25--32.}
    \label{fig:routing_individual_sources}
\end{subfigure}
\caption{\textbf{\attnres{} routing-source composition} (ten-clip mean). In (a), dashed lines mark block resets; the first attention layer carries only the input source.}
\label{fig:routing_source_diagnostics}
\end{figure}

%% file: figure_latex/routing_rank_diagnostics.tex
\begin{figure}[htbp]
\centering
\begin{subfigure}[b]{0.49\textwidth}
    \centering
    \includegraphics[width=\linewidth]{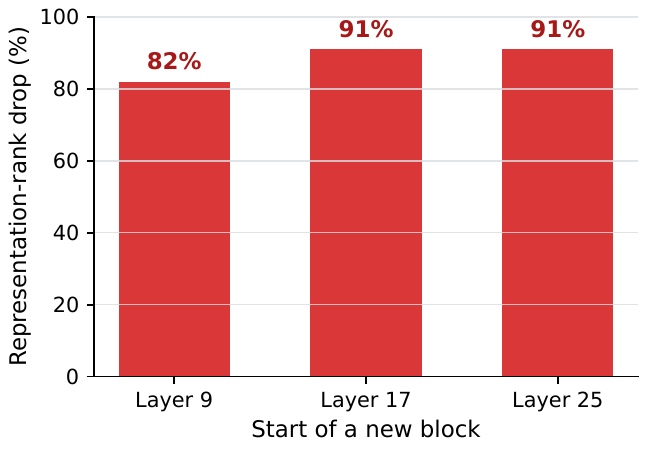}
    \caption{Earlier-block feature removal.}
    \label{fig:rank_ablation}
\end{subfigure}\hfill
\begin{subfigure}[b]{0.49\textwidth}
    \centering
    \includegraphics[width=\linewidth]{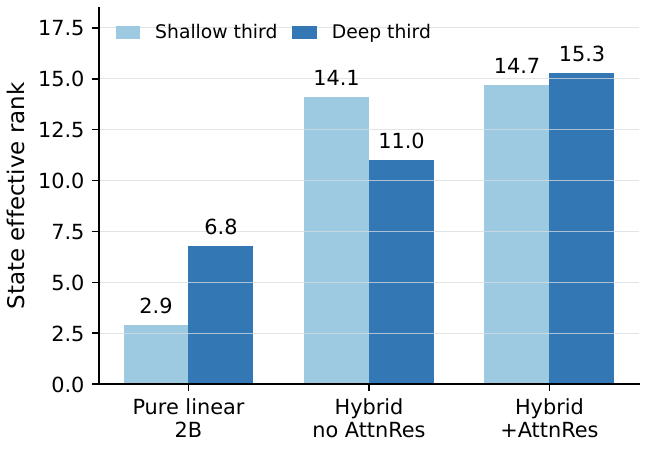}
    \caption{Step-unmatched rank context.}
    \label{fig:rank_pure_linear}
\end{subfigure}
\caption{\looseness=-1 \textbf{Routing intervention and rank context.} (a) Representation-rank drop after removing completed-block sources at block entries. (b) Descriptive comparison across separately trained models that differ in width, training horizon, and checkpoint.}
\label{fig:routing_rank_diagnostics}
\end{figure}

%% file: figure_latex/hardware_scaling_14b.tex
\begin{figure}[htbp]
    \centering
    \includegraphics[width=\textwidth]{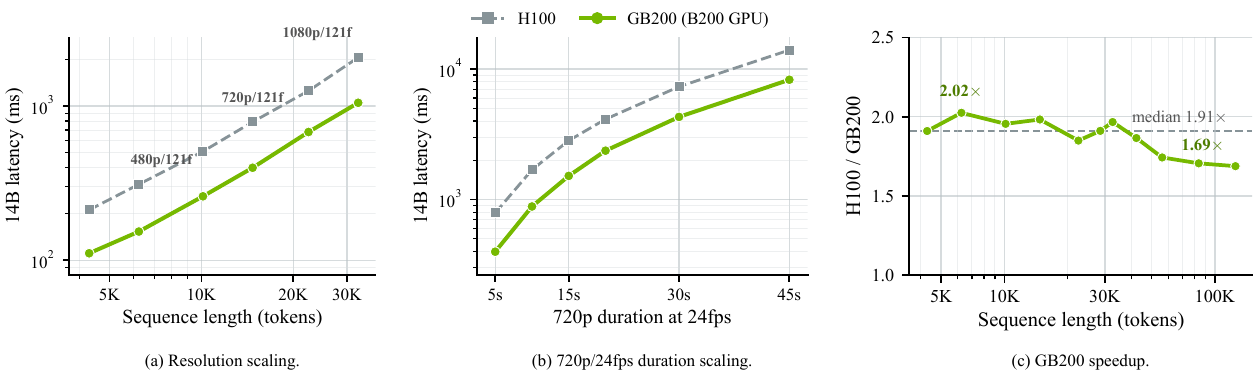}
    \caption{\looseness=-1 \textbf{Matched 40-layer 14B cross-hardware DiT-forward profiling} (width 4,096; one GPU, bf16, batch one; \attnres{} off). (a) Resolution/frame scaling. (b) 720p duration scaling. (c) H100/GB200 latency ratio. Both devices use identical inputs and compiled kernels through 125.1K tokens.}
    \label{fig:hardware_scaling_14b}
\end{figure}

%% file: figure_latex/architecture_scaling_14b.tex
\begin{figure}[htbp]
    \centering
    \includegraphics[width=\textwidth]{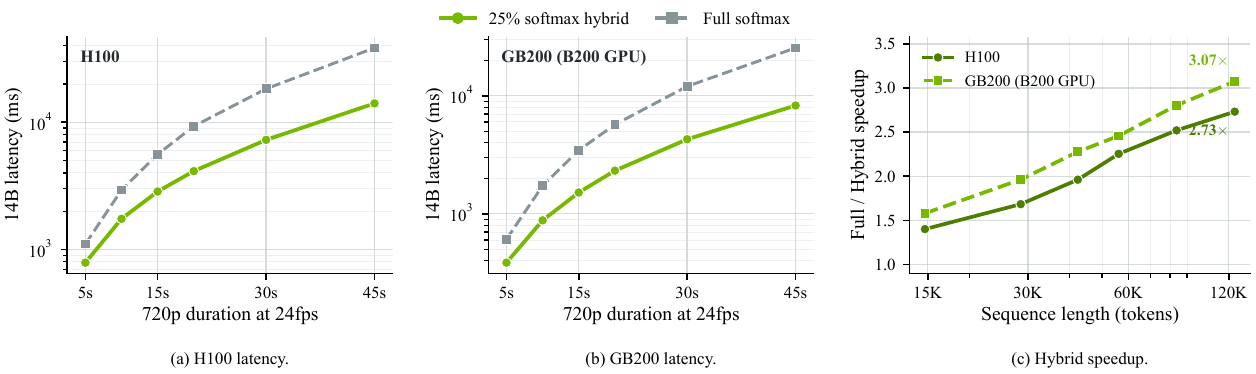}
    \caption{\looseness=-1 \textbf{Matched 40-layer 14B hybrid versus full-softmax DiT-forward profiling} (width 4,096; one GPU, bf16, batch one; \attnres{} off). (a,b) Latency over the 720p/24fps duration grid on H100 and GB200. (c) Full/Hybrid latency ratio. Both layouts use the same width, depth, inputs, Flash SDPA implementation for softmax, and compiled protocol.}
    \label{fig:architecture_scaling_14b}
\end{figure}

%% file: figure_latex/temporal_conv_runtime.tex
\begin{figure}[t]
\centering
\begin{minipage}[t]{0.62\linewidth}
    \vspace{0pt}
    \input{table_latex/temporal_conv_scaling}
\end{minipage}\hfill
\begin{minipage}[t]{0.35\linewidth}
    \vspace{0pt}
    \input{figure_latex/runtime_composition}
\end{minipage}
\end{figure}

%% file: table_latex/temporal_conv_scaling.tex
\centering\small
\captionof{table}{\textbf{Controlled FFN latency swap} at 720p (one H100; ms). $\Delta$ is temporal-convolution overhead; $\dagger$ marks the excluded 60s pair.}
\label{tab:temporal_conv_scaling}
\setlength{\tabcolsep}{2.5pt}
\renewcommand{\arraystretch}{1.38}
\begin{tabular}{l l rrrrrr}
\toprule
Backbone & FFN & 5s & 15s & 20s & 30s & 45s & 60s \\
\midrule
\multirow{3}{*}{$L{=}20,d{=}2240$} & SwiGLU & 52 & 126 & 161 & 242 & 367 & 510 \\
 & Temporal-conv & 64 & 157 & 202 & 302 & 455 & 627 \\
 & $\Delta$ & +12 & +31 & +42 & +60 & +88 & +117 \\
\midrule
\multirow{3}{*}{$L{=}32,d{=}2560$} & SwiGLU & 306 & 906 & 1258 & 2036 & 3416 & 4994$^\dagger$ \\
 & Temporal-conv & 396 & 1141 & 1573 & 2499 & 4091 & 10043$^\dagger$ \\
 & $\Delta$ & +90 & +236 & +315 & +463 & +677 & ---$^\dagger$ \\
\bottomrule
\end{tabular}

%% file: figure_latex/runtime_composition.tex
\centering
\captionof{figure}{\textbf{Eager module composition} at 22.1K/111.3K tokens under the controlled forward-profile protocol.}
\label{fig:runtime_composition}
\vspace{-0.8em}
\includegraphics[width=\linewidth]{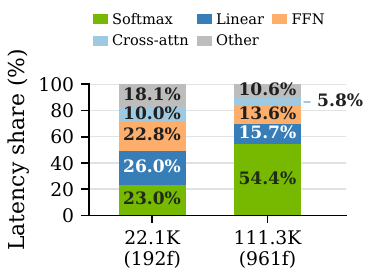}

%% file: figure_latex/appendix_samples.tex
\begin{figure}[p]
    \centering
    \includegraphics[width=\textwidth]{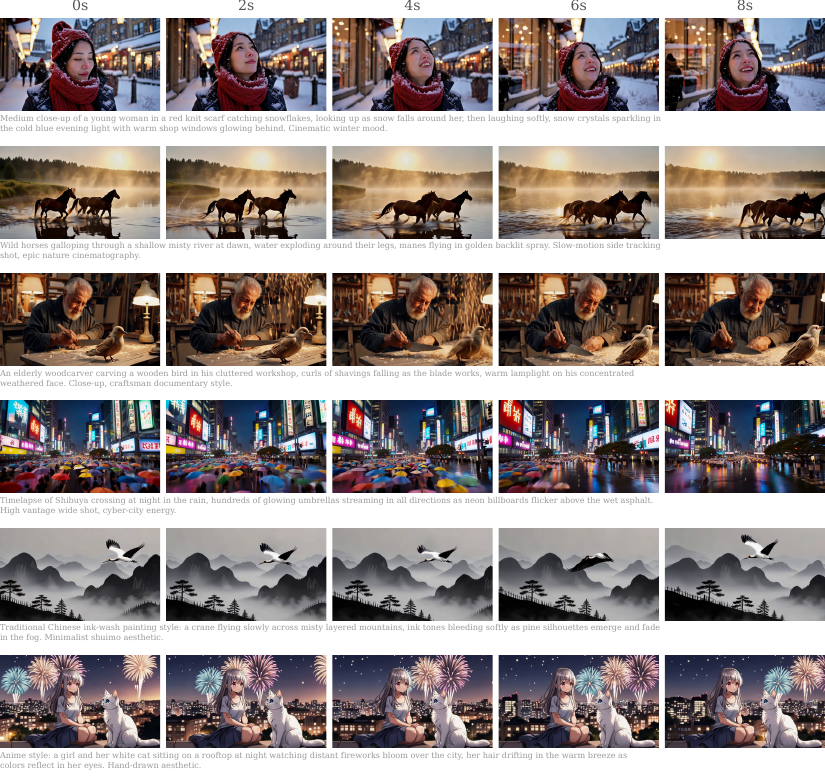}
    \caption{\textbf{Additional 720p text-to-video samples.} Each row unrolls one 8s $1280{\times}736$ clip as five frames at $0/2/4/6/8$s; the generating prompt is shown beneath each row.}
    \label{fig:more_samples}
\end{figure}

%% file: figure_latex/online_rl_qualitative.tex
\begin{figure}[p]
    \centering
    \includegraphics[width=\textwidth,height=0.91\textheight,keepaspectratio]{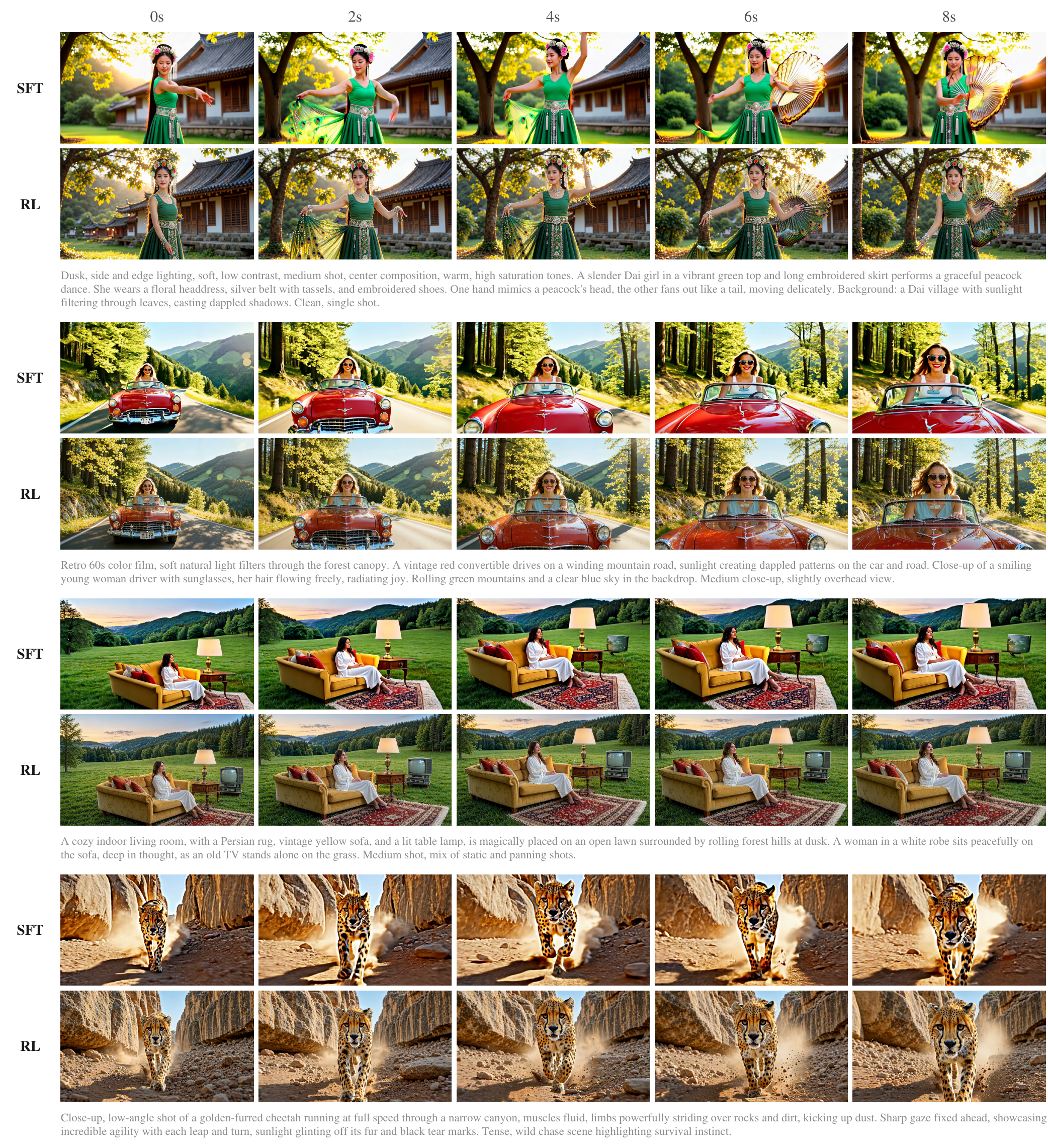}
    \caption{\looseness=-1 \textbf{Qualitative effect of online RL across four prompts.} Each block compares generations from the SFT and RL checkpoints using the same prompt and classifier-free guidance scale (CFG 8). Columns show frames at $0/2/4/6/8$s, and the complete generating prompt is reproduced beneath each pair.}
    \label{fig:online_rl_qualitative}
\end{figure}

%% file: figure_latex/method_compare_t2v.tex
\begin{figure}[p]
    \centering
    \includegraphics[width=\textwidth,height=0.94\textheight,keepaspectratio]{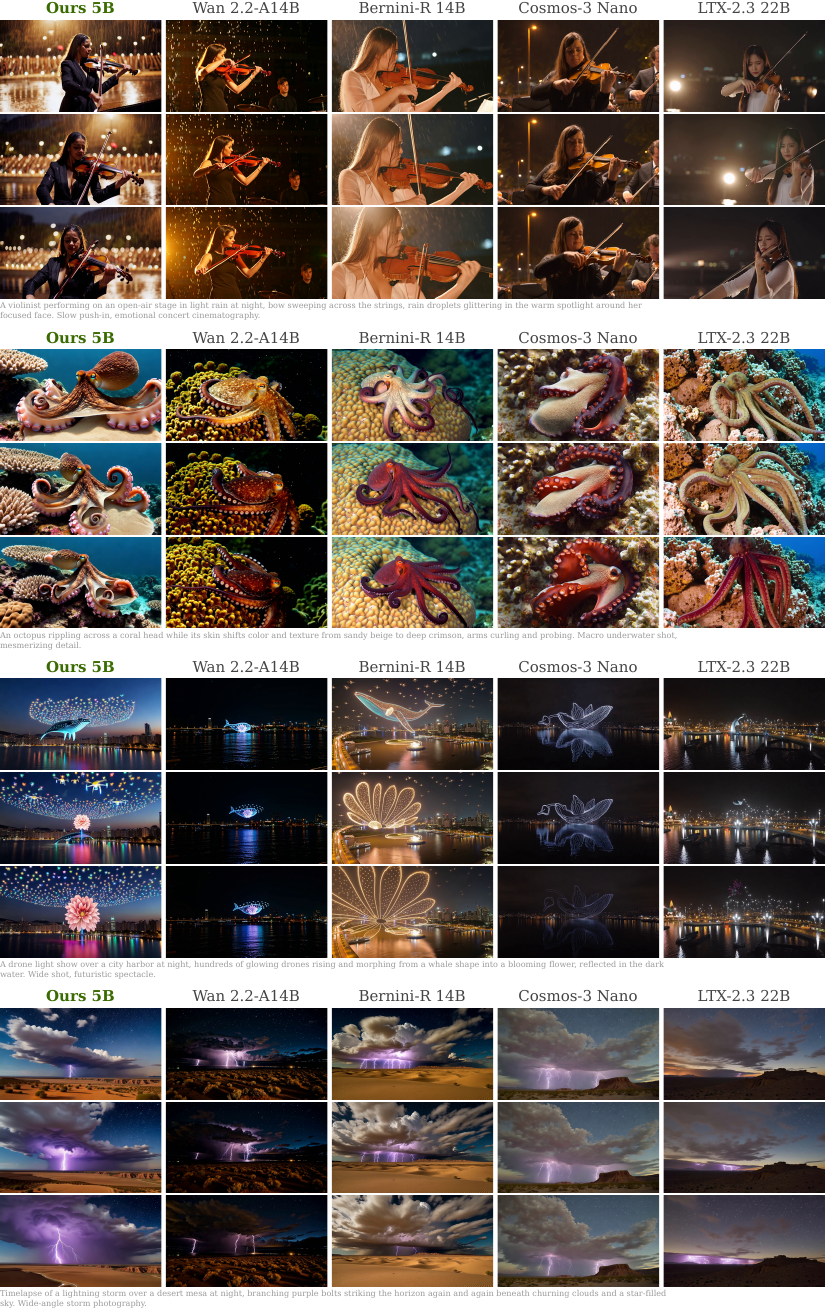}
    \caption{\textbf{Cross-method 720p text-to-video comparison} on four held-out prompts. Columns are methods and rows show $0/50/100\%$ of each clip; prompts appear below the corresponding blocks.}
    \label{fig:t2v_compare}
\end{figure}

%% file: figure_latex/ti2v_compare_a.tex
\begin{figure}[p]
    \centering
    \includegraphics[width=\textwidth]{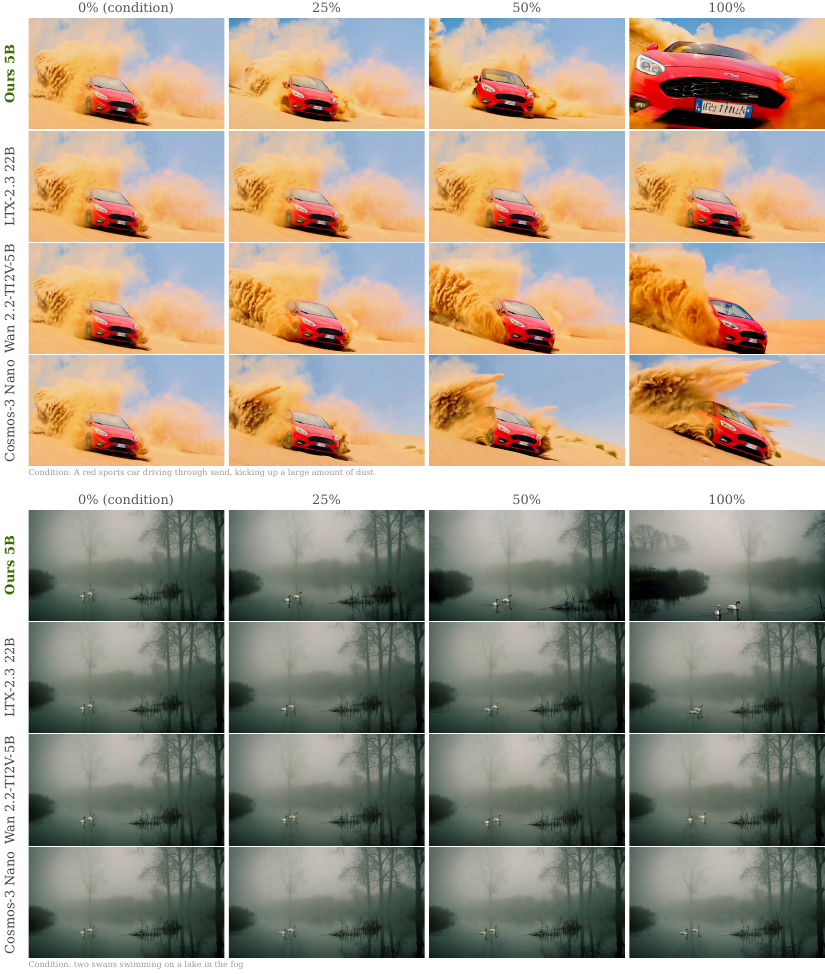}
    \caption{\looseness=-1 \textbf{First-frame-conditioned video comparison, pairs 1--2.} Each row shows one method at $0/25/50/100\%$ of its clip; the first frame is shared. Protocol details are in Appendix~\ref{app:method_compare}.}
    \label{fig:ti2v_compare_a}
\end{figure}

%% file: figure_latex/ti2v_compare_b.tex
\begin{figure}[p]
    \centering
    \includegraphics[width=\textwidth]{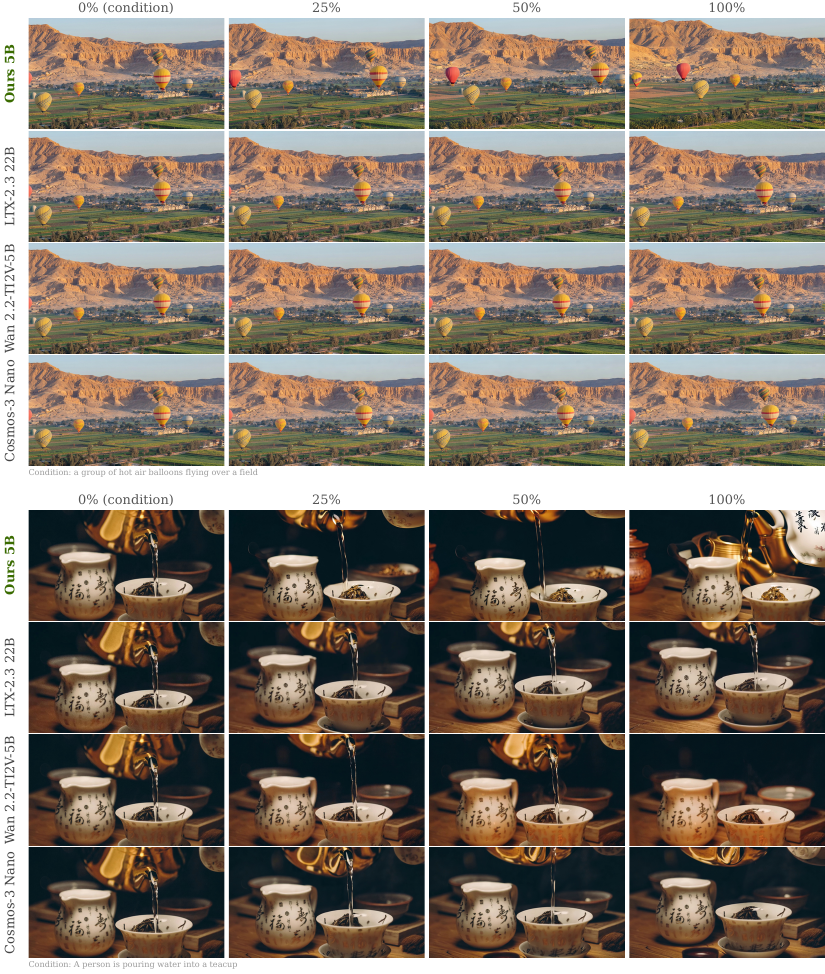}
    \caption{\looseness=-1 \textbf{First-frame-conditioned video comparison, pairs 3--4.} The protocol matches Figure~\ref{fig:ti2v_compare_a}.}
    \label{fig:ti2v_compare_b}
\end{figure}